\documentclass[authoryear, baselineskip, 3p, 11pt]{elsarticle}
\usepackage{amsfonts, amsmath, bm, amsthm, amssymb}
\usepackage{nicefrac}
\usepackage{graphicx}
\usepackage{xcolor}
\usepackage{caption}
\usepackage{subcaption}
\usepackage{comment}

\usepackage{pdflscape}

\usepackage{parskip} 
\usepackage{setspace}
\usepackage{booktabs, multirow, makecell}
\usepackage{adjustbox}

\usepackage[ruled]{algorithm2e}

\newcommand{\ra}[1]{\renewcommand{\arraystretch}{#1}} 
\newcommand{\w}[1]{\widetilde{#1}}
\DeclareMathOperator*{\argmax}{argmax}

\usepackage{natbib}
\usepackage[font=scriptsize]{caption}
\usepackage[hyphens]{url} 
\usepackage{hyperref}
\hypersetup{
    urlcolor  = black,
	colorlinks = true,
	citecolor = blue, 
	linkcolor = blue 
}

\begin{document}

\title{Disambiguation of Company names via Deep Recurrent Networks}
\author[1]{Alessandro~Basile\fnref{dr}}
\ead{alessandro.basile@intesasanpaolo.com}

\author[1]{Riccardo~Crupi\fnref{dr}}
\ead{riccardo.crupi@intesasanpaolo.com}

\author[2]{Michele~Grasso}
\ead{michele.grasso@earlymorning.com}

\author[1]{Alessandro~Mercanti\fnref{dr}}
\ead{alessandro.mercanti@intesasanpaolo.com}

\author[1]{Daniele~Regoli\fnref{dr}}
\ead{daniele.regoli@intesasanpaolo.com}

\author[1]{Simone~Scarsi\fnref{dr}}
\ead{simone.scarsi@intesasanpaolo.com}

\author[1]{Shuyi~Yang\fnref{dr}}
\ead{shuyi.yang@intesasanpaolo.com}

\author[1]{Andrea~Claudio~Cosentini\fnref{dr}}
\ead{andrea.cosentini@intesasanpaolo.com}

\address[1]{Data Science \& Artificial Intelligence, Intesa Sanpaolo S.p.A., Italy}

\address[2]{Early Morning S.r.l., Italy}


\fntext[dr]{The views and opinions expressed are those of the authors and do not necessarily reflect the views of Intesa Sanpaolo, its affiliates or its employees.}

\begin{abstract}
Name Entity Disambiguation is the Natural Language Processing task of identifying textual records corresponding to the same Named Entity, i.e. real-world entities represented as a list of attributes (names, places, organisations, etc.). In this work, we face the task of disambiguating companies on the basis of their written names. 
We propose a Siamese LSTM Network approach to extract --- via supervised learning --- an embedding of company name strings in a (relatively) low dimensional vector space and use this representation to identify pairs of company names that actually represent the same company (i.e. the same Entity). 

Given that the manual labelling of string pairs is a rather onerous task, we analyse how an Active Learning approach to prioritise the samples to be labelled leads to a more efficient overall learning pipeline. 

With empirical investigations, we show that our proposed Siamese Network outperforms several benchmark approaches based on standard string matching algorithms when enough labelled data are available. Moreover, we show that Active Learning prioritisation is indeed helpful when labelling resources are limited, and let the learning models reach the out-of-sample performance saturation with less labelled data with respect to standard (random) data labelling approaches.

\end{abstract}

    
\maketitle

\section{Introduction}

A common information retrieval task with several applications is the association of company names from any internal or external source to a specific company registered in an internal database.
A straightforward method based on the equality of strings is insufficient to tackle the issue, since company names are not uniquely spelt and abbreviations or synonyms vary across databases. 
Often, data sources contain company names manually entered, which may include typos, abbreviations or indeed mistakes of any kind.
 
Some exemplar applications include:
\begin{enumerate}
    \item integrating data from other sources into the internal database to supplement companies' information.
    \item Linking the name of a company mentioned in the news or the social media to an actual company in an internal registry so that it may be associated with, e.g. sentiment or ESG scores.
    \item The sender and receiver of a bank transfer are in general manually entered, and do not need to match specific registry (contrary to the transfer code). Nevertheless, it is useful to correctly identify the sender and receiver companies, e.g. to detect fraudulent behaviours.
\end{enumerate}

Company name disambiguation refers to the task of determining whether two given strings encoding company names actually represent the same company. For example, the two strings ``Intesa Sanpaolo S.p.A.'' and ``Intesa San Paolo bank'' in fact represent the same corporation. This task is an instance of the so-called Named Entity Disambiguation (NED) \citep{aghaebrahimian2020named}, whose aim is to identify data records corresponding to the same Named Entity, i.e. real-world entities represented as a list of attributes (names, places, organisations, etc.).

It must be said that the NED task has, ironically, a somewhat fuzzy meaning in the literature, and it is used to refer to more or less different problems. Moreover, many other labels are often employed in the literature to refer to slightly similar --- when not identical --- tasks. We refer to \cite{barlaug2021neural}, who try to introduce a somewhat general framework, whose specific instances can be matched to various NED-related tasks and sub-tasks.     

In particular, Entity Disambiguation, or Entity Matching, should not be confused with Entity Recognition, which is the task of mapping a string (usually a word or a group of words inside a longer text sequence) to a finite set of classes (the Entities), without --- in general --- the explicit need of \emph{comparing} strings~\citep{kolitsas2018end}. 

The contribution of this work is twofold: on the one hand, we propose a Siamese Recurrent Neural Network approach to the ask of disambiguating pairs of company names. 
We show, via experiments, that the proposed approach outperforms other baseline models and that is efficient in generalising to other domains. On the other hand, we use our proposed model in an Active Learning setting to demonstrate how to make human labelling more efficient by prioritising the samples to be labelled.

The rest of the paper is organised as follows: Section~\ref{sec:related_works} is devoted to a discussion of relevant literature, in particular regarding NED and Active Learning. In Section~\ref{sec:methods} we detail the methodologies we use for the NED task: we describe both our proposed model and the baseline approaches we use as benchmark. Section~\ref{sec:al} is devoted to describing how we implement the Active Learning setting. In Section~\ref{sec:data} we thoroughly describe how we build the datasets that we use in the experiments. The latter are described in Section~\ref{sec:experiments}. Discussion of the insights derived from the experimental results is presented in Section~\ref{sec:discussion}, while Section~\ref{sec:conclusion} contains concluding remarks. 

The python code implementation of the Siamese Neural Network model, of the Active Learning setting and of all the experiments, is available in open-source at~\href{https://github.com/rcrupiISP/SiameseDisambiguation}{github.com/rcrupi/SiameseDisambiguation}.

\section{Related Works}
\label{sec:related_works}

\subsection{Entity Disambiguation}

Most classical approaches to strings pair matching leverage string similarity measures, quantifying how much two given strings are similar with a more or less sophisticate deterministic rule~\citep{cohen2003comparison, christen2006comparison, sun2015comparative}. These methods are built on several background encodings of given strings, such as phonetic-based, character-based, or based on term frequency/inverse term frequency (\emph{tf-idf}) and hybrid versions of these~\citep{christen2006comparison}. Other approaches try to employ also semantic knowledge to compute similarity~\citep{prasetya2018performance}.


More recently, methodologies based on \emph{learning the appropriate similarity function} from a sample of data of the desired domain have become quite common. 
For instance, \cite{piskorski2020tf} employ \emph{tf-idf} vectors of $n$-grams as predictors for a Machine Learning (ML) classifier. The internal (learned) representation can then be exploited as an abstract vector summarising the information relevant to the task. Finally, a standard vector similarity function --- such as cosine similarity --- of the representations of two strings can be used to infer their task-specific distance. 

In the domain of toponym matching --- i.e. pairing of different strings representing the same real-world location --- \cite{santos2018toponym} have faced a problem very similar to the one we are discussing. They propose an approach based on a \emph{Siamese} Deep Neural Network architecture~\citep{chicco2021siamese} and benchmark it against several distance-based methodologies and several classifiers taking as input various pairwise string distances. They conclude that their approach is superior in terms of matching performance, even if less efficient in terms of computational time with respect to pure distance methods.

\cite{neculoiu2016learning} propose a Siamese Deep Learning model as well, with the slightly different task of mapping strings to a predefined set of job names. This classification task is nevertheless approached by translating it into a NED framework and by learning a vector representation of strings, such that close vectors correspond to the same job class. They use two different loss terms for positive and negative matches: in particular, the loss term for negative samples is zero below a certain threshold, so as not to pay a cost for non-matching pairs that are increasingly dissimilar.
Furthermore, they introduce some interesting data augmentation techniques, such as observations derived by adding some random typos and character-level deletion in positive pair samples.

\cite{aghaebrahimian2020named} also propose a Deep Learning approach to NED, while having as input, not the raw strings, but rather character-level \emph{tf-idf} vectors of $n$-grams, similarly to~\cite{piskorski2020tf}. Moreover, they train the network in a contrastive fashion, i.e. by feeding input triplets with the observed string and both an actual string match (i.e. string representing the same entity --- positive example) and a non-matching string (negative example).

We refer to \cite{barlaug2021neural} for a thorough review of Neural Networks-based approaches to the NED task.

\subsection{Active Learning}

Entity Matching datasets are typically constructed through laborious human labelling. 
To increase the efficiency of such a procedure, techniques have been proposed to carefully prioritise the samples to be labelled. Against this background, \emph{Active Learning} --- i.e. the sub-field of ML with the characteristic that the learning algorithm is allowed to choose the data from which it learns~\citep{settles2009active, arora2007active} --- has been proven to be beneficial in the Entity Matching domain~\citep{meduri2020comprehensive}.
In particular, the selection criteria of candidate observations to be labelled are usually expressed in terms of \emph{informativeness} and \emph{representativeness} \citep{zhou2018brief}. While representativeness-based approaches try to find a pattern of the unlabelled data, using graphs or clustering methods, informativeness-based approaches --- such as Uncertainty sampling and query-by-committee --- choose the instances to be labelled based on how uncertain they are to be classified. In particular, Query-by-Committee approaches propose to train several classifiers and define the uncertainty of a given observation based on the rate of disagreement on their predictions. They are sometimes referred to as the learner-agnostic approaches. Uncertainty sampling, on the other hand, given a specific classifier, employs the distance from the decision boundary as a proxy for uncertainty. It is therefore referred to as learner-aware approach.

While in \cite{meduri2020comprehensive} the task is to disambiguate two instances on the basis of several different information sets (e.g. address, name, etc.), in this work we focus on company disambiguation based on string names only. A disambiguation task involving couples of words (e.g. {`principle', `principal'}, and {`end', `and'}) was faced in \cite{banko2001scaling}.
In particular, they extract features from the words and apply a ML classifier to estimate their similarity. Active Learning (specifically query-by-committee) is then exploited to iteratively select batches from a pool of unlabelled samples. Half of the samples in each batch are selected randomly, while the other half are selected on the basis of their uncertainty.

Since Deep Learning requires huge amount of data, Active Learning is particularly well suited to limiting the data labelling but keeping high the performance~\citep{zhan2022comparative}. In particular, in this work we adopt a modification of the ``Least confidence'' approach as query strategy~\citep{huang2021deepal}. 

Other works, such as \cite{sorscher2022beyond}, suggest a self-supervised data pruning method. In contrast to Active Learning, data reduction is done in a single step. The self-supervised metric is based on the application of $k$-means over the embedding space of a pre-trained network: an instance far away from its cluster centroid is considered uncertain. Experiments in \cite{sorscher2022beyond} suggest that the instances to be removed actually depend on the size of the starting dataset: if it is large enough, it is beneficial to include the most uncertain samples, whereas if it is small, it is preferable to include simplest (least uncertain) samples.

\section{Methods}
\label{sec:methods}
In ML settings, a multiclass classification function $\mathcal{C}: X \mapsto Y$  takes as input a feature vector $\mathbf{x} \in X$ from the input feature space $X$ and outputs a class label $y \in Y$ from a finite set of possible $\Bar{\alpha}$ classes $Y = \{0, 1, 2, \ldots, \Bar{\alpha}-1\}$ \citep{hastie2009elements}. 
Most families of ML classifiers actually learn to estimate the probabilities
\begin{equation}
    \mathbb{P}(y=\alpha \mid \mathbf{x}),\quad \alpha \in \left\{0, 1, \ldots, \bar{\alpha}\right\}.
\end{equation}
In the following, we label with $\hat{y}_{i, \alpha}$ the estimated probability that the $i$-th sample belongs to the class~$\alpha$. Being probabilities, the $\hat{y}_{i, \alpha}$ are such that $\sum_{\alpha=0}^{\bar{\alpha}-1} \hat{y}_{i, \alpha}=1$ and $\hat{y}_{i, \alpha} \in [0, 1] \; \forall \alpha \in Y$.
The predicted class $\Bar{y}_i$ is the one associated with the highest probability, namely
\[
    \bar{y}_i = \argmax_{\alpha \in Y} \hat{y}_{i,\alpha}.
\]
The case with only two class labels ($\bar{\alpha}=2$) --- i.e. binary classification tasks --- are usually formalised as: 
\begin{equation}\label{eq:binaryclass}
    \mathcal{C}: X \mapsto [0, 1],
\end{equation}
where the output $\hat{y}_i$ is an estimate of $\mathbb{P}(y_i=1 \mid \mathbf{x}_i)$ and corresponds to the output $(\hat{y}_{i,0}, \hat{y}_{i,1}) =(1-\hat{y}_i, \hat{y}_i)$ in the generic multiclass setting.

In this work, we frame the string match problem as a binary classification task, where the components of the input feature vector $\mathbf{x}_i$ are couples of strings $\mathbf{x}_i = \{a_i, b_i\}$, and the classifier estimates the probability $\hat{y}_i$ that the two strings correspond to the same company (we label 1 the matching class).
Calling $\mathbf{S}$ the set of possible character symbols, we may formalise the string matching classifier as a function
\begin{equation}\label{eq:stringclass}
    \mathcal{C}: \mathbf{S}^n \times \mathbf{S}^n \mapsto [0, 1],
\end{equation}
where the integer $n$ denotes the fixed (maximum) length of the strings to be analysed.

\subsection{Baseline methods}\label{sec:baseline}
To determine how dissimilar two strings $a$ and $b \in \mathbf{S}^n$ are, in the following we shall make use either of a distance function --- $\mathcal{D}(a, b) \in \mathbb{R}^{+}$ where 0 stands for identical strings and the higher the distance the more dissimilar $a$ and $b$ --- or a similarity function --- $\mathcal{S}(a, b) \in [0,1]$, 1 denoting identical strings. 

\subsubsection*{Levenshtein}
Widely used deterministic methods are the \emph{edit distance} metrics. Generally speaking, they are based on counting the number of operations needed to transport a string onto another. The choice of the type of operations allowed determines the specific form of the distance. The most widely known edit-distance metric is the \emph{Levenshtein distance} (sometimes referred to as \emph{the} edit-distance), which calculates the distance as the number of insertions, deletions, and substitutions required to transform one string into another. The formula of the Levenshtein distance $\mathcal{D}_\text{Lev}(a, b)$ between two strings $a=a_{0}a_{1}a_{2}\dots a_{m}$ of length $m$ and $b=b_{0}b_{1}b_{2}\dots b_{n}$ of length $n$ is given by the following recursion:
\begin{equation}
    \mathcal{D}_\text{Lev}(a, b)=\left\{\begin{aligned}
        &m &\text{ if } n=0,\\
        &n &\text{ if } m=0,\\
        &\mathcal{D}_{Lev}(\w{a}, \w{b}) &\text{ if } a_{0}=b_{0},\\
        &1+min\left\{\begin{matrix}
                \mathcal{D}_\text{Lev}(\w{a}, b) \\
                \mathcal{D}_\text{Lev}(a, \w{b})\\
                \mathcal{D}_\text{Lev}(\w{a}, \w{b})\\
                \end{matrix}\right. &\text{ otherwise};
        \end{aligned}\right.
\end{equation}
where $\w{x}$ denotes the string $x$ without the first character ($x_{0}$), i.e. $\w{x}=x_{1}x_{2}\dots x_{s}$.

A closely related edit-based distance is the \emph{InDel distance}, which allows only insertions and deletions. It is easy to see that the InDel distance is equivalent to the Levenshtein distance where the substitution operation is assigned a cost of 2 (deletion + insertion). 
We call \emph{InDel Ratio} the following normalised version of the InDel distance:
\begin{equation}\label{eq:InDel_ratio}
    \mathcal{R}_\text{ID}(a, b) = \left(1 - \frac{\mathcal{D}_{ID}(a, b)}{\lvert a \rvert + \lvert b \rvert}\right),
\end{equation}
where $\mathcal{D}_\text{ID}(a, b)$ denotes the InDel distance between $a$ and $b$. 
We make use of the Python library \href{https://github.com/seatgeek/thefuzz}{\texttt{TheFuzz}} to compute $\mathcal{R}_\text{ID}(a, b)$, with the sole difference that \texttt{TheFuzz} expresses the ratio in percentage points.

\subsubsection*{Jaro-Winkler similarity}
The \emph{Jaro-Winkler (JW) similarity} ($\mathcal{S}_\text{JW}$) is another edit-based metric emphasising both the amount of matching characters and their placement inside the two strings. Notice that this is a notion of similarity, i.e. $\mathcal{S}_\text{JW} \in [0, 1]$, $\mathcal{S}_\text{JW}(a, b) = 0$ corresponding to no match at all between two strings $a$ and $b$, and $\mathcal{S}_\text{JW}(a,b) = 1$ to exact match. 

The Jaro-Winkler similarity is a variant of the Jaro similarity, that, for two strings $a$ and $b$, is defined as
\begin{equation}
    \mathcal{S}_\text{J}(a, b) = \left\{\begin{aligned}
        &0 &\text{ if } c=0,\\
        &\frac{1}{3}\left(\frac{c}{\lvert a \rvert} + \frac{c}{\lvert b \rvert}+\frac{c-t}{c}\right) &\text{ otherwise},
        \end{aligned}\right.
\end{equation}
with:
\begin{itemize}
    \item $c$ the number of matching characters. Two characters are considered a match when they are the same and they are no more than $\frac{\max\left(\lvert a \rvert, \lvert b \rvert\right)}{2}-1$ chars apart of one another,
    \item $t$ is the number of transpositions counted as the number of matching characters found in the wrong order, divided by two.
\end{itemize}
The JW~similarity extends the definition of the Jaro similarity by favouring strings with a matching prefix:
\begin{equation}
    \mathcal{S}_\text{JW}(a, b) = \mathcal{S}_\text{J}(a, b) + \ell\, p\left(1 - \mathcal{S}_\text{J}(a, b)\right),
\end{equation}
where $\ell$ is the length of the common prefix up to a maximum value, and $p$ is a constant scaling factor determining the strength of the premium. The maximum value attributed to $\ell$ and the value of $p$ should be chosen such that $\ell p \leq 1$. 

\subsubsection*{Jaccard similarity}
While the edit-based metrics look at what is necessary to do to transform one string into another, the \emph{token-based} metrics consider the strings as sets of tokens (i.e. the words or characters composing the strings) and search for the common tokens between two sets. A widely-used token-based similarity metric is the \emph{Jaccard similarity} which is defined as the ratio of the intersection over the union of the token's sets $A=\{a_{0}, a_{1}, a_{2}, \dots, a_{m}\}$ and $B=\{b_{0}, b_{1}, b_{2}, \dots, b_{m}\}$ for the two strings $a$ and $b$ respectively (sometimes referred to as IoU --- Intersection over Union), i.e.
\begin{equation}
    \mathcal{S}_\text{Jac}=\frac{\lvert A \cap B\rvert}{\lvert A \cup B\rvert}.
    \label{eq:jaccard_sim}
\end{equation}
Notice that the Jaccard metric does not take into account the order of tokens, unlike the previously discussed edit-based metrics.

\subsubsection*{Token Set Ratio}
Another approach for string matching is represented by the \emph{approximate string matching} algorithms \citep{navarro2001guided}
that leverage the basic notions of distance introduced so far, but take into account matching also at substring level.
In particular, we make use of the so-called \emph{Token Set Ratio} metric computed via \texttt{TheFuzz} Python library. It works as follows:

\begin{enumerate}
    \item takes the \emph{unique words} (i.e. substrings separated by whitespaces) for each string $a$ and $b$, let us call them $W_a$ and $W_b$, respectively:
    \item builds the following word sets: 
    \begin{align*}
        & I_{ab} = W_a \cap W_b,\\ 
        & W_{a\setminus b} = W_a \setminus W_b,\\
        & W_{b\setminus a} = W_b \setminus W_a;
    \end{align*}
    \item sorts alphabetically the sets and builds new strings $s_{ab}$, $s_{a\setminus b}$, $s_{b\setminus a}$ by joining with whitespaces the words in the corresponding (sorted) sets; 
    \item builds the new strings: $c_a$ joining $s_{ab}$ and $s_{a\setminus b}$ with a whitespace, and analogously $c_b$ with $s_{ab}$ and $s_{b\setminus a}$;
    \item compute the similarity as
    \begin{equation}
        \mathcal{R}_\text{TS}(a, b) = \max\left\{
        \begin{aligned}
            \mathcal{R}_\text{ID}(s_{ab}, c_a)\\
            \mathcal{R}_\text{ID}(s_{ab}, c_b)\\
            \mathcal{R}_\text{ID}(c_a, c_b)
        \end{aligned}\right. . 
    \end{equation}
\end{enumerate}

\subsubsection*{Baseline classifier}
To build a classifier based on the match algorithms just described, the strings are pre-processed by removing punctuation and capitalising the text.
\begin{table}[ht]
    \centering
    \ra{1.3}
    \begin{tabular}{@{}lcr@{}}
      \toprule
      \textbf{score} & \textbf{type} & \textbf{range}\\
      \midrule
      Levenshtein & distance & $\mathcal{D}_\text{Lev}(a, b) \in [0, {\rm max}\left(\lvert a\rvert, \lvert b\rvert\right)]$\\
      InDel & similarity & $\mathcal{R}_\text{ID}(a, b) \in [0, 1]$\\
      Jaro-Winkler & similarity & $\mathcal{S}_\text{JW}(a, b) \in [0, 1]$\\
      Token Set Ratio & similarity & $\mathcal{R}_\text{TS}(a, b) \in [0, 1]$ \\
      Jaccard & similarity & $\mathcal{S}_\text{Jac}(a, b) \in [0, 1]$\\
      \bottomrule
    \end{tabular}
    \caption{List of metrics used as features in the Baseline and Random Forest classifier.}
    \label{tab:features_in_RF}
\end{table}
The five string match algorithms listed in Table~\ref{tab:features_in_RF} constitute our baseline methods and
for each of the 5 string matching score listed in Table~\ref{tab:features_in_RF}, we do the following:
\begin{itemize}
    \item pre-process the strings with the cleaning method,
    \item applies the selected string match algorithm to each pair of strings in the training dataset,
    \item train a Decision Stump --- i.e. a Decision Tree with a single node --- given as input the score just computed, and as label the match/non-match nature of each pair of strings.
\end{itemize}

\subsection{Random Forest classifier}
\label{sec:RF}
The validity of string similarity algorithms presented so far depends on the use case. Therefore we use a \emph{Random~Forest} classifier which uses the five string match metrics listed in Tab. \ref{tab:features_in_RF} as input features at the same time.

The Random Forest pipeline goes as follows:
\begin{itemize}
    \item pre-process the strings with the same cleaning method as for the Baseline Trees,
    \item for each pair of strings in the training dataset, compute the 5 scores listed in Table~\ref{tab:features_in_RF},
    \item extract 2 additional features from each string: the \textit{number of characters} and the \textit{number of words} (i.e. substrings split with respect to whitespaces),
    \item train a Random Forest classifier with 9 features in input (5 matching scores + 2 $\times$ number of words + 2 $\times$ number of characters), and as label the match/non-match nature of each pair of strings.
\end{itemize}

The popular \texttt{Scikit-Learn} Python library~\citep{scikit-learn} is used both for the Decision Stump of the Baseline classifiers and for the Random Forest implementations.  The hyperparameters of the Random Forest are set to: {\tt max\_depth=3, n\_estimators=100, class\_weight=`balanced'}.

\subsection{Proposed Approach}
\label{sec:siamese}

We propose an approach based on Recurrent Neural Networks (RNNs) employing a Siamese strategy~\citep{bromley1993signature}, framing the learning problem as a binary classification of string pairs.

To keep the format of the input consistent, each string is preprocessed as follows: the strings are padded to a length of 300 chars\footnote{The longest string in our whole dataset has 124 chars. If longer strings are to be fed to the model, then a truncation to 300 is implemented.}, using the heavy division sign as placeholder for padding. The string cleaning is in this case limited to the uppercase. The rationale is to leave as much information as possible to the Neural Network model to learn useful patterns. 
Each string is then tokenised character-wise and one-hot encoded with an alphabet of 63 symbols\footnote{Another possible approach is using the entire word as a single token. Experiments using this approach resulted in poor performances, because the model cannot properly handle cases of spelling errors or abbreviations} (i.e. the placeholder plus $62$ symbols for capital letters, numbers, punctuation, and whitespace), resulting in a $300\times 63$ input matrix.

\begin{figure}
    \centering
    \includegraphics[width=.45\textwidth]{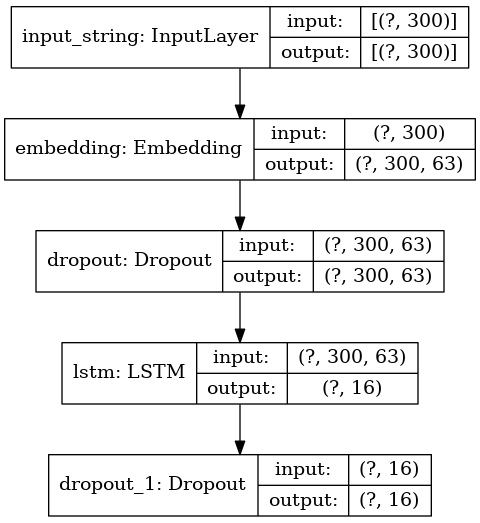}
    \caption{\textbf{Embedding model}. Schematic {\tt TensorFlow} representation of the Embedding model described in Section~\ref{sec:siamese}. Each block denotes a {\tt TensorFlow} layer, with input and output tensor dimensions. As usual in {\tt TensorFlow} representations, the generic batch size is denoted with the symbol `?'.}
    \label{fig:embedding_architecture}   
\end{figure}

Each input matrix, representing a string, is then processed by an Embedding Model (Figure~\ref{fig:embedding_architecture}), composed of a $300 \times 63$ embedding matrix (i.e. a matrix whose entries are learned via loss optimisation) followed by an LSTM layer with 16 nodes. Weights sharing is employed during learning between the encoding model of the two strings to force the two encoding models to be identical (which is indeed the origin of the name ``siamese''), in such a way as to preserve the symmetry of the problem and to effectively learn a unique representation space for individual strings. Notice that we don't make use of the full sequence of hidden LSTM states (that would be a $300 \times 16$ matrix) as output of the LSTM layer --- i.e. the embedding representation of individual strings --- we instead employ the hidden state corresponding to the last token in the string, that nevertheless implicitly contains information of all the hidden states along the sequence --- this is indeed the main feature of RNN models\footnote{We experimented the same architecture with a Bidirectional LSTM layer in place of a plain LSTM layer, but without any sign of improvement in performance, while on the other hand, the computational cost increased significantly.}.

\begin{figure*}
\begin{center}
    \includegraphics[width=.9\textwidth]{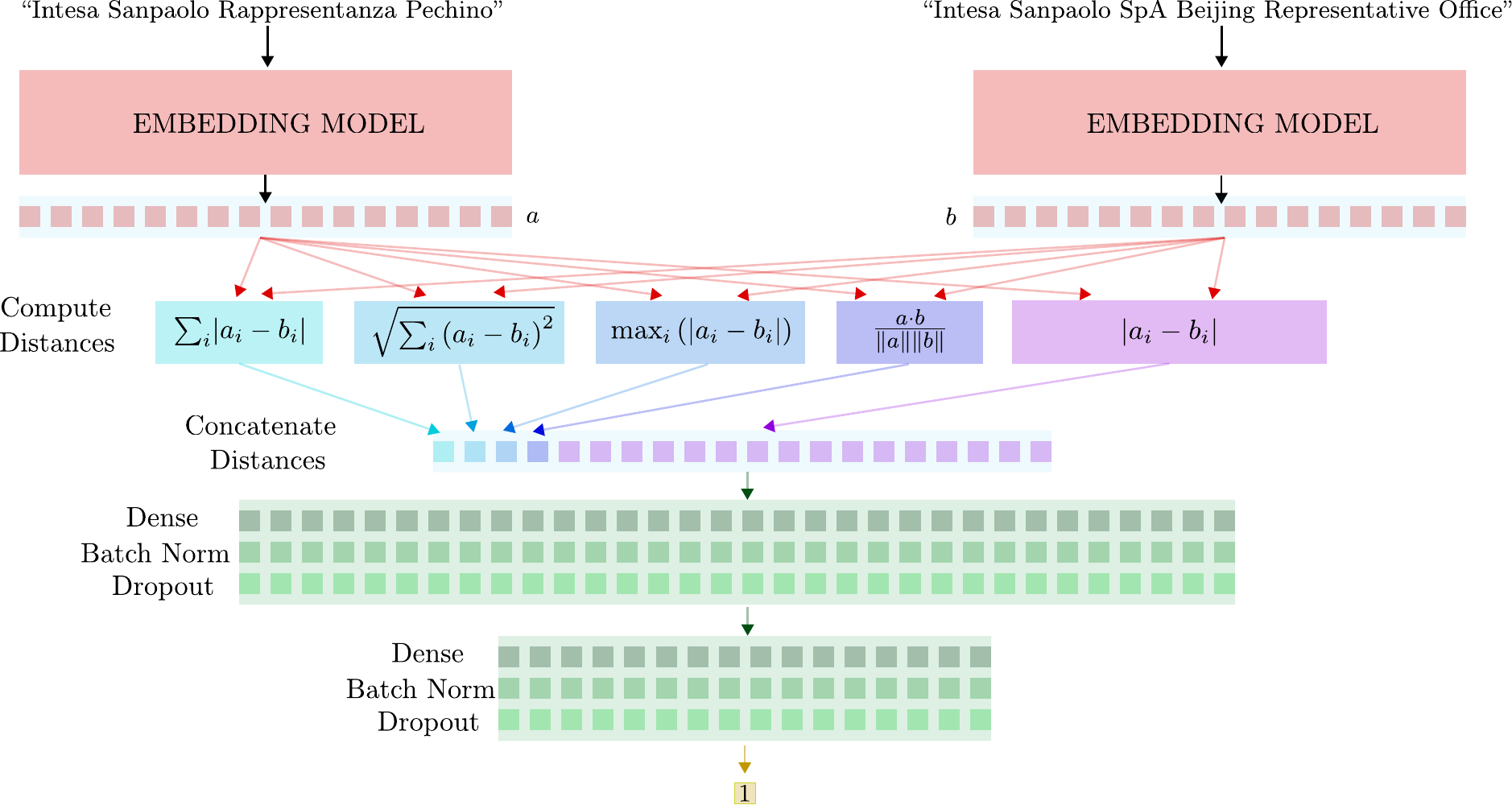}
    \caption{\textbf{Siamese architecture}. Two input strings are processed by the same LSTM-based encoding model (see Figure~\ref{fig:embedding_architecture}) to get a 16-dim vector representation each. These are used to compute several distances: ${L_1}$, ${L_2}$, ${L_{\infty}}$, cosine and the element-wise absolute difference. This information is then concatenated --- obtaining a 20-dim vector --- and fed to two consecutive blocks, each composed of a dense layer with ReLU activations, batch normalisation, and dropout. The first block has a dense layer with 32 nodes, while the second block has a dense layer with 16 nodes. The final layer is a single neuron with a sigmoid activation function.} 
    \label{fig:siamese_model_architecture}   
\end{center}
\end{figure*}

The two encodings thus generated are then employed to compute several vector distances --- namely, ${L_1}$, ${L_2}$, ${L_{\infty}}$, cosine-based distance and the element-wise absolute difference. These results are then fed as inputs to a Feed-Forward Neural Network classifier, composed of two consecutive blocks, each consisting of a dense layer with ReLU activations, batch normalisation, and dropout. The final output is a single neuron with a sigmoid activation function, to get the classification score (Figure~\ref{fig:siamese_model_architecture}).

Since every operation performed on the inputs has the commutative property, the whole model $\mathcal{C}^s$, composed by the ensamble of the encoding model and the prediction model, has commutative property, then given any two input strings $a$ and $b$, we have $\mathcal{C}^s(a,b) = \mathcal{C}^s(b,a) $ by design.
Binary Crossentropy is used as loss function, as usual for binary classification tasks. We employ Nadam as optimisation algorithm, with parameter choice $\beta_1=0.8$, $\beta_2=0.9$ and a fixed learning rate $\epsilon=10^{-4}$.
Both the Embedding model and the downstream Feed-Forward classifier are implemented in Python via {\tt TensorFlow}~\citep{tensorflow2015-whitepaper}.

The rationale for employing a Siamese approach is that of learning a high-level embedding of strings, where the similarity in the embedding space reflects the probability of being instances of the same entity.

\section{Active Learning}
\label{sec:al}

\begin{figure*}
    \centering
    \includegraphics[width=.9\textwidth]{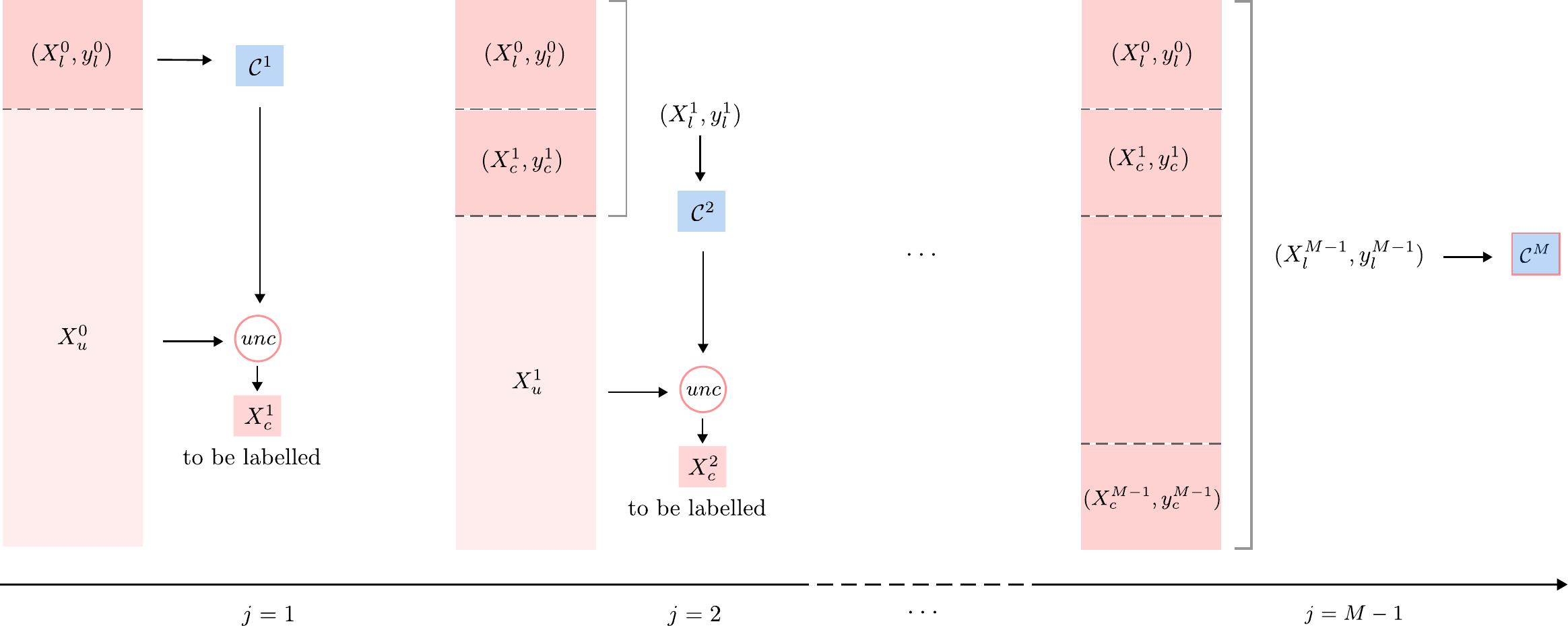}
    \caption{\textbf{Active Learning}. Illustrative representation of the Active Learning procedure outlined in Algorithm~\ref{alg:activelearning}.}
    \label{fig:al}
\end{figure*}

In an ideal scenario, a predictive model can be built from labelled data in a fully supervised way: generally speaking, increasing the amount of labelled data improves the generalisation capacity of the learned model. However, in a real-world scenario, the number of labelled instances is often limited: the labelling process is often costly, time-consuming and oftentimes requires high-level domain knowledge. On the other hand, unlabelled data are in general much easier to collect and may be used to improve the predictive performance of the model. 

Formally, in a classification setting, there are:
\begin{itemize}
    \item a set of labelled instances $(X_l, y_l)$ (where $X_l$ represents features and $y_l$ denotes the corresponding labels); 
    \item a set of not-labelled-yet instances $X_u$. 
\end{itemize}

If the labelling process is cost-effective, one could get the labels $y_u$ of not-labelled-yet instances $X_u$, and then use a fully-supervised learning algorithm~\citep{sen2020supervised}. In a completely opposite situation, where no additional labels can be collected, or at a prohibitive cost, semi-supervised methods have been proved to be effective~\citep{DBLP:journals/ml/EngelenH20}. 

Oftentimes, the situation is in the middle: given the available resources, only a limited number of instances can be labelled. How to effectively exploit these labelling resources is the focus of Active Learning~\citep{settles2009active, aggarwal2014active, DBLP:journals/csur/RenXCHLGCW22}: given the limited labelling capability, how to choose the subset $X_c \subset X_u$ to label in order to obtain the maximum performance gain? In practice, $X_c$ is chosen and constructed according to some query strategies in a multi-step procedure. 

In our work, we adopt an uncertainty sampling strategy \citep{settles2009active} where, at each step, instances of $X_u$ on which the prediction of the most updated model is less certain are selected. These data points are removed from $X_u$, labelled by domain experts and then added to $(X_l, y_l)$ in order to train an improved version of the classifier, with the rationale that, since the additional data points are picked near the decision boundary instead of being randomly selected, they contain more valuable information for the model learning.

Let $x_i \in X_u$ be an unlabelled instance, we denote with $(\hat{y}_{i,0}, \ldots, \hat{y}_{i,\bar{\alpha}-1})$ the probabilities estimated by the classifier over the $\bar{\alpha}$ classes, such that $\sum_\alpha \hat{y}_{i, \alpha} = 1$. Then, we can define the uncertainty as 

\begin{equation}
\label{eq:unc}
    unc(\hat{y}_{i}) = 1 - \max_\alpha \hat{y}_{i, \alpha} .
\end{equation}

In the case of a binary classification setting ($\bar{\alpha}=2$), it is equivalent to measuring the distance of the positive class predicted probability from $\frac{1}{2}$:

\begin{equation}\label{eq:unc_bin}
    unc(\hat{y}_i) = \frac{1}{2} - \left\lvert \frac{1}{2} - \hat{y}_{i} \right\rvert .
\end{equation}

At each step of the training process, instances of $X_u$ are sorted according to the uncertainty defined above and the top $B$ most uncertain samples are added to $X_l$ and removed from $X_u$. A classifier is then trained on the updated version of $X_l$.  

Empirical evidence during experiments suggests that using directly the uncertainty defined in Equation~\eqref{eq:unc} is sub-optimal in balancing the exploration and exploitation threshold~\citep{banko2001scaling}: 
feeding the learner with only uncertain most samples, especially at the beginning, could result in a batch of data too biased towards difficult instances, leading to poor generalisation capability. To prevent this, and alternatively to the strategy introduced by \cite{banko2001scaling} --- i.e. using batches composed of half of samples selected randomly and the other half selected on the basis of uncertainty ---  we propose to use the following noisy version of uncertainty:

\begin{equation}
    unc_{\sigma}(\hat{y}_i) = unc(\hat{y}_i) + \epsilon_i, \quad \epsilon_i \sim \mathcal{N}(0, \sigma),
    \label{eq:unc_noise}
\end{equation}

where $\epsilon_i$ are independent and identically distributed normal random variables and $\sigma$ denotes the level of the noise we would like to introduce. The random noise introduced in Equation~\eqref{eq:unc_noise} helps the learner to generalise better during the first stages of the Active Learning procedure. 

In Algorithm~\ref{alg:activelearning} we formalise the entire procedure described above, while Figure~\ref{fig:al} shows an illustrative representation.

\begin{algorithm}
\footnotesize
\caption{Active Learning Procedure}
\label{alg:activelearning}

\SetKwInOut{Input}{input}\SetKwInOut{Output}{output}
\Input{$X_{l}^0$ (initial labelled instances)\\ 
$y_{l}^0$ (labels relative to $X_{l}^0$)\\
$X_{u}^0$ (initial unlabelled instances)\\ 
$M$ (number of iterations)\\
$B^1, B^2, \ldots, B^{M-1}$ (batch sizes)\\
$\sigma$ (noise for uncertainty)}
\Output{a trained classifier $\mathcal{C}^M$}

\BlankLine
Train an initial classifier $\mathcal{C}^1$ on $(X_{l}^0, y_{l}^0)$ \\
\For{$j = 1, \ldots, M-1$}{
    \tcp{Predict the probabilities over the instances, according to equation \eqref{eq:stringclass}}
    $\{ \hat{y} = \mathcal{C}^j(x) \mid x \in X_u^{j-1} \}$ \\
    \tcp{Compute $unc_\sigma$ for each point in $X_u^{j-1}$ according to equations \eqref{eq:unc_bin} and \eqref{eq:unc_noise}}
    $\left\{unc_\sigma(\hat{y}) = \frac{1}{2} - \left\lvert \frac{1}{2} - \hat{y} \right\rvert  + \epsilon \right\}$\\
    \tcp{Select from $X_u^{j-1}$ the $B^j$ instances with highest $unc_\sigma$}
    $X_{c}^j$ $\leftarrow$ top $B^j$ instances with respect to $unc_\sigma$\\
    \tcp{Label the selected instances}
    $y_{c}^j$ $\leftarrow$ domain expert labels relative to $X_{c}^j$\\
    \tcp{Update the set of labelled instances} 
    $X^j_l \leftarrow X_l^{j-1} \cup X_{c}^j$\\
    $y_l^{j} \leftarrow y_l^{j-1} \cup y_{c}^j$\\
    \tcp{Update the set of unlabelled instances} 
    $X_u^{j} \leftarrow X_u^{j-1} \setminus X_{c}^j$\\
    Train a classifier $\mathcal{C}^{j+1}$ on $(X_{l}^{j}, y_{l}^{j})$\\
}
\KwResult{$\mathcal{C}^M$}
\BlankLine

\end{algorithm}

\section{Data}
\label{sec:data}

The data employed in our analysis are extracted from two different domains, i.e. company registry and bank transfers. More specifically, the data consist of couples of strings being: 
\begin{enumerate}
    \item the company names (concatenated with the address) of the same entity as recorded in two different datasets obtained by external data providers.
    \item the beneficiary names of the same entity as recorded in SWIFT\footnote{\href{https://www.swift.com/}{Society for Worldwide Interbank Financial Telecommunications}.} bank transfers.
\end{enumerate}
The two sources are used independently, with the first used for training and testing and the second only for testing.

\subsection{Data labelling}
\label{sec:data_labelling}
Each instance of the datasets used in the experiments consists of a pair of names and a binary target variable: we use the label 1 when the pair of names correspond to the same company, and label 0 otherwise. 
As stated in previous sections, the labelling process is time-consuming since it involves the identification of raw data usually with human annotation. To tackle this task, we adopt a 2-step strategy: we pre-label some couples of names with a rule-based criterion suggesting the target variable (match/non-match), and then we check the suggested labels manually. 
The rule to pre-label company string pairs is based on the domain of data we are considering.

\subsubsection*{Pre-labelling for company registry data}
Many companies are identified through the Legal Entity Identifier (LEI)
: aliases with the same LEI 
refer to the same company entity.
We leverage this background to identify candidate couples with positive labels (same LEI) or with negative labels (different LEI). At the end of this process, these suggested labels are manually validated. 
We label 9,000 couples of names from the company registry data (with a 1:4 positive/negative label ratio). 
Figure~\ref{fig:3fold_distribution} displays the distribution of JW~similarity relative to the 9,000 couples conditioned on the label matching. 

It is worth noting that the pre-labelling strategy based on the LEI code goes beyond the simple string similarity between company names. Indeed, the LEI code can be used to identify named entities even when they undergo various types of legal transactions, such as mergers, acquisitions, consolidations, purchases and management acquisitions. In this case, the company name can vary after the legal transaction, while still referring to the same entity. Of course, these \emph{counter-intuitive positive matches} are beyond the range of validity of the methods we are discussing in this work --- based solely on the similarity of string names --- but we decided to include some of them to test their limits. We discuss some of these examples in Section~\ref{sec:discussion}.

\subsubsection*{Pre-labelling strategies for bank transfer}
In a bank transfer, funds are transferred from the bank account of one entity (the sender or payer) to another bank account (of the beneficiary or recipient). Beneficiaries with the same bank account (IBAN) refer to the same company and can be used to identify candidate couples with a positive label. More challenging is the construction of couples with negative labels, identified by recipients with different IBAN. The reason is that the same company may own more than one IBAN, and this requires a more detailed validation. 
With this strategy, we label 200 couples from the bank transfers (with a 1:1 positive/negative labels ratio).

We keep these data separate from the previously discussed 9,000 pairs, and we use them only for testing purposes, with the rationale of verifying the robustness of our approaches under domain shift scenarios (see Section~\ref{sec:training_and_test_sets}).

\subsection{Training and test sets}
\label{sec:training_and_test_sets}

We adopt a stratified $k$-fold approach to split the data: we split the 9,000 labelled instances into $3$ subsets $S_1$, $S_2$, and $S_3$ with equal size and taking care to maintain the same positive/negative ratio. At each iteration, we choose one of them as the test set $S_{\text{test}}$ and keep the union of the rest as a training set $S^\text{L}_{\text{train}}$.

\begin{figure}
    \centering
    \begin{subfigure}{.45\textwidth}
        \includegraphics[width=\linewidth]{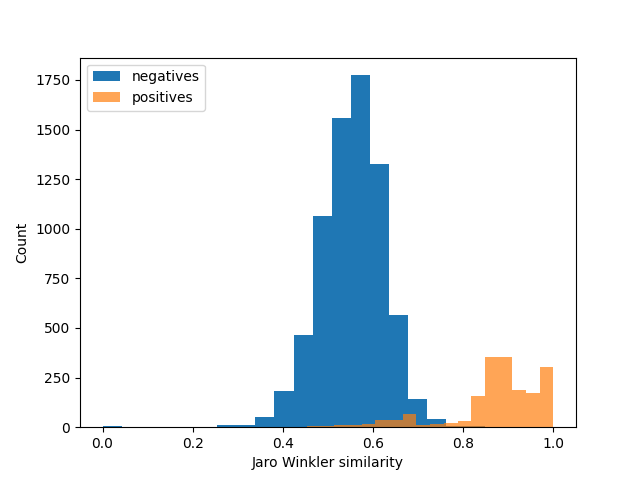}
        \caption{Distribution of similarity over all string pairs.}
        \label{fig:3fold_distribution}        
        \end{subfigure}%
    \begin{subfigure}{.45\textwidth}
        \centering
        \includegraphics[width=\linewidth]{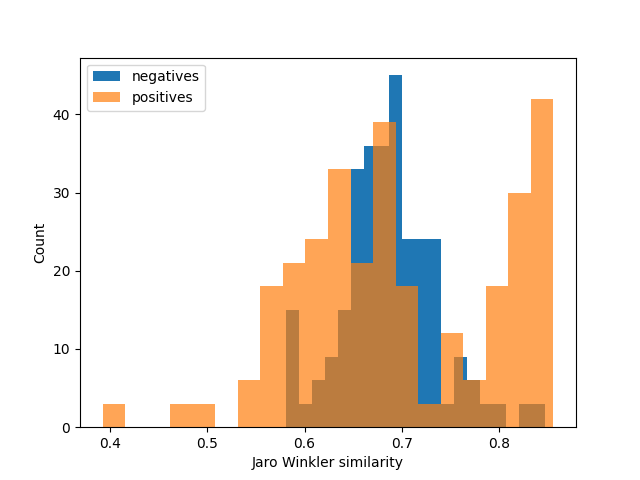}
        \caption{Distribution of similarity on JW-ordered test set.
        }
        \label{fig:difficult_test_distribution}           
    \end{subfigure}
    \caption{\textbf{Similarity distribution over string pairs}. Distribution of JW~similarity conditioned on the label matching over (\subref{fig:3fold_distribution}) the 9,000 labelled instances and (\subref{fig:difficult_test_distribution}) JW-ordered test set. The continuous score is binned with size $0.05$. In (\subref{fig:3fold_distribution}) the histogram shows that the negative (i.e. non-matching) samples are roughly normally distributed around 0.55, while the majority of positive (i.e. matching) samples have JW~similarity between $0.8$ and $1$. In (\subref{fig:difficult_test_distribution}), on the contrary, the distributions of positive and negative pairs are largely overlapping, making it difficult to determine a threshold between the two classes, in particular with respect to the whole dataset.}
    \label{fig:distributions}
\end{figure}

From $S_{\text{train}}^\text{L}$ we sample $\tfrac{1}{3}$ of its elements to form a \emph{medium size training set} $S_{\text{train}}^\text{M}$, and then we again sample $\tfrac{1}{20}$ of instances contained in $S_{\text{train}}^\text{M}$ to obtain a \emph{small size training set} $S_{\text{train}}^\text{S}$. Therefore, we end with 3 training sets for each fold: $S_{\text{train}}^\text{S} \subset S_{\text{train}}^\text{M} \subset S_{\text{train}}^\text{L}$. 
These 3 training sets with increasing size are useful to analyse how the different approaches perform in relation to the amount of data they are given to learn from (see experiments described in Section~\ref{sec:experiments} and the corresponding discussion in Section~\ref{sec:discussion}).

Analogously, we define different test sets: the \emph{randomly ordered test set} $S_{\text{test}}^\text{RO}=S_{\text{test}}$ (i.e. the entire hold-out set available), and the \emph{JW ordered test set} $S_{\text{test}}^\text{JO}$ obtained by ranking $S_{\text{test}}^\text{RO}$ instances according to their \emph{JW~similarity}\footnote{As reported in the experiments (see Table~\ref{tab:results}), the JW method performs better compared to the other baseline methodologies. Therefore, selecting test instances based on this metric is a way to check the robustness of ML approaches.} and taking the top 100 negative cases (i.e. non-matching pairs that are nevertheless mostly JW-similar) and the bottom 100 positive cases (i.e. matching pairs that are nevertheless mostly JW-dissimilar). The distribution of JW~similarity of samples in $S_{\text{test}}^\text{JO}$ conditioned on matching labels is shown in Figure~\ref{fig:difficult_test_distribution}.

To test the robustness of the methods presented in Section~\ref{sec:methods}, we introduce a third test set $S_{\text{test}}^\text{DS}$ in addition to $S_{\text{test}}^\text{RO}$ and $S_{\text{test}}^\text{JO}$, extracted from a different data source. Namely, as anticipated at the beginning of Section~\ref{sec:data}, we extract pairs of company names from SWIFT bank transfer registry, where transaction payers write the beneficiary without any form of oversight. The dataset is balanced, with $100$ positive examples obtained by matching recipients with the same IBAN and $100$ negative cases obtained by random matching. We name this dataset the \emph{domain shifted test set}.

We expect the $S_{\text{test}}^\text{JO}$ and $S_{\text{test}}^\text{DS}$ test sets to be particularly challenging for the string matching algorithms given in Section~\ref{sec:methods}: indeed $S_{\text{test}}^\text{JO}$ is, by design, a stress test for the single-feature based methods, and it can be used to estimate how effectively the Random Forest and the Siamese Network are able to generalise with respect to the baselines. The $S_{\text{test}}^\text{DS}$ dataset instead is likely to be challenging for several reasons: it is drawn from a different dataset with respect to the training set; not only the beneficiary names may be affected by typos and all sorts of noise due to the free writing, but they are also not in a standardised form, i.e. they may contain additional information such as the company address\footnote{This motivates the use of names and address in the data extracted from company registry data.}.

\section{Experiments}\label{sec:experiments}
Different modelling strategies (string distance metrics, Deep Neural Networks, Active Learning) have different (dis)advantages.
In order to compare them fairly and point out the best scenario in which to apply each of them, we prepare two different experimental settings: 
\begin{enumerate}
    \item standard supervised classification setting,
    \item Active Learning setting. 
\end{enumerate}
In each of these two settings, we run the experiments employing a 3-fold cross-validation strategy as described in Section~\ref{sec:training_and_test_sets}. 

\subsection{Supervised classification setting}

We evaluate how the performances of the models presented in section \ref{sec:methods} change as they are trained on training sets of different sizes ($S_{\text{train}}^\text{L}$, $S_{\text{train}}^\text{M}$ and $S_{\text{train}}^\text{S}$).  Each of the training sets is used to train each of the following $7$ models:
\begin{itemize}
    \item $5$ different Baseline Trees (see Section~\ref{sec:methods}), based on Levenstein distance ($\mathcal{D}_\text{Lev}$), InDel ratio ($\mathcal{R}_\text{ID}$), Token Set Ratio ($\mathcal{R}_\text{TS}$), Jaccard similarity ($\mathcal{S}_\text{Jac}$), and JW~similarity ($\mathcal{S}_\text{JW}$);
    \item a Random Forest classifier, introduced in Section~\ref{sec:RF};
    \item our proposed Siamese Network, introduced in Section~\ref{sec:siamese}. 
\end{itemize}
Out-of-sample performance is evaluated on test sets $S_{\text{test}}^\text{RO}$, $S_{\text{test}}^\text{JO}$, and $S_{\text{test}}^\text{DS}$ by computing the Balanced Accuracy (BA), thus giving equal weights to positive and negative classes, irrespective of actual class imbalances. More precisely, as argued in~\cite{chicco2021matthews}, BA is a good measure --- preferable over, e.g. Matthews Correlation Coefficient (MCC) and $F_1$ score --- when the aim is to compare classifiers across datasets with different class imbalances, and/or when the focus is to correctly classify the ground truth instances, which is exactly what we are doing.

Table~\ref{tab:results} summarises experimental results discussed in Section~\ref{sec:discussion}. 
We leave to Table~\ref{tab:complete_results} in the appendix additional metrics computed for the experiments ($F_1$~score and MCC).


\begin{table*}
\centering
\ra{1.3}
\resizebox{\textwidth}{!}{
\begin{tabular}{@{}l c c c c c c c c c c@{}}
\toprule
\makecell[l]{\textbf{training}\\ \textbf{set size}} & \makecell[c]{\textbf{test set}\\ \textbf{type}} & \phantom{ab} & \textbf{Levenshtein} & \makecell[c]{\textbf{InDel}\\ \textbf{Ratio}} & \makecell[c]{\textbf{Token} \\\textbf{Set Ratio}} & \textbf{Jaccard} & \textbf{JW} & \makecell[c]{\textbf{Random}\\ \textbf{Forest}} & \makecell[c]{\textbf{Siamese}\\ \textbf{Network}}\\

\midrule

\multirow{3}{*}{small}  & \textit{RO} && $0.665 \pm 0.045$ & $0.855 \pm 0.025$ & $0.935 \pm 0.005$ & $0.577 \pm 0.09$ &$\mathbf{ 0.957 \pm 0.003}$ & $0.951 \pm 0.017$ & $0.892 \pm 0.042$ \\
 & \textit{JO} && $0.428 \pm 0.061$ & $0.505 \pm 0.005$ & $0.662 \pm 0.006$ & $0.438 \pm 0.06$ & $0.678 \pm 0.016$ & $0.678 \pm 0.018$ & $\mathbf{ 0.725 \pm 0.044}$ \\
 & \textit{DS} && $0.582 \pm 0.05$ & $0.717 \pm 0.01$ & $0.723 \pm 0.015$ & $0.533 \pm 0.058$ & $0.717 \pm 0.003$ & $0.718 \pm 0.02$ & $\mathbf{ 0.735 \pm 0.013}$ \\
[5pt] 
\multirow{3}{*}{medium}  & \textit{RO} && $0.643 \pm 0.014$ & $0.878 \pm 0.013$ & $0.944 \pm 0.007$ & $0.523 \pm 0.041$ & $0.957 \pm 0.003$ & $0.965 \pm 0.006$ & $\mathbf{ 0.975 \pm 0.002}$ \\
 & \textit{JO} && $0.473 \pm 0.038$ & $0.488 \pm 0.029$ & $0.652 \pm 0.003$ & $0.463 \pm 0.064$ & $0.675 \pm 0.013$ & $0.697 \pm 0.013$ & $\mathbf{ 0.867 \pm 0.019}$ \\
 & \textit{DS} && $0.535 \pm 0.009$ & $0.735 \pm 0.009$ & $0.743 \pm 0.003$ & $0.5 \pm 0.0$ & $0.715 \pm 0.0$ & $0.743 \pm 0.008$ & $\mathbf{0.773 \pm 0.016}$ \\
[5pt] 
\multirow{3}{*}{large}  & \textit{RO} && $0.641 \pm 0.012$ & $0.871 \pm 0.014$ & $0.944 \pm 0.007$ & $0.523 \pm 0.04$ & $0.956 \pm 0.001$ & $0.967 \pm 0.004$ & $\mathbf{ 0.976 \pm 0.002}$ \\
 & \textit{JO} && $0.495 \pm 0.0$ & $0.5 \pm 0.0$ & $0.652 \pm 0.003$ & $0.465 \pm 0.061$ & $0.687 \pm 0.006$ & $0.72 \pm 0.023$ & $\mathbf{ 0.903 \pm 0.051}$ \\
 & \textit{DS} && $0.53 \pm 0.0$ & $0.733 \pm 0.012$ & $0.743 \pm 0.003$ & $0.5 \pm 0.0$ & $0.715 \pm 0.0$ & $0.743 \pm 0.003$ & $\mathbf{ 0.777 \pm 0.01}$ \\

\bottomrule
\end{tabular}
}
\caption{\textbf{Experimental results}. Balanced Accuracy of the Random Forest, Siamese Neural Network, and the 5 single-distance Decision Trees, all trained with datasets of different sizes (small, medium, large) and tested on the three test sets discussed in Section~\ref{sec:training_and_test_sets}, namely randomly ordered ($S_\text{test}^\text{RO}$), JW-ordered ($S_\text{test}^\text{JO}$) and domain shifted ($S_\text{test}^\text{DS}$). Mean and standard deviation computed via stratified $k$-fold cross-validation are displayed. Contrary to the 5 baseline models, the performances of the Random~Forest and the Siamese Neural Network improve with the size of the training set. Moreover, the Siamese Neural Network trained on medium and large datasets outperforms the other approaches. Bold figures denote row-wise maximum values.}
\label{tab:results}
\end{table*}


\subsection{Active Learning setting}
\label{sec:al_setting}

We then employ an Active Learning strategy --- outlined in Section~\ref{sec:al} and Algorithm~\ref{alg:activelearning} --- to train both the Random Forest and the Siamese Neural Network. 
In our experiments, the initial labelled instances $(X^{0}_{l}, y^{0}_{l})$ consist of 100 samples and correspond to $S_{\text{train}}^\text{S}$, while $X_u^0$ consista of the residual $5900$ couples, namely $S_{\text{train}}^\text{L} \setminus S_{\text{train}}^\text{S}$. We fix the $\sigma$ parameter at $1/6$ for all the experiments. 
The batch sizes are set in such a way that all instances are spanned with $M = 9$ iterations. More formally, for $j=1, 2,\ldots, M-1$, the batch size at the $j$-th iteration is
\begin{equation*}
    B^j=\left\{\begin{aligned}
                & 100 \times 2^{j-1} &  j \in [1, 4], \\
                & 800 &  j = {5, 6}, \\
                & 1400 &  j > 6.
                \end{aligned}
                \right.
\end{equation*}
This choice is motivated by the idea of better tracking the impact of the Active Learning approach: indeed, we expect the greater benefits to come in the very first phases, while the marginal benefit after having seen enough data will be negligible.   

\begin{figure}[ht!]
    \centering
    \begin{subfigure}{.45\textwidth}
        \includegraphics[width=\linewidth]{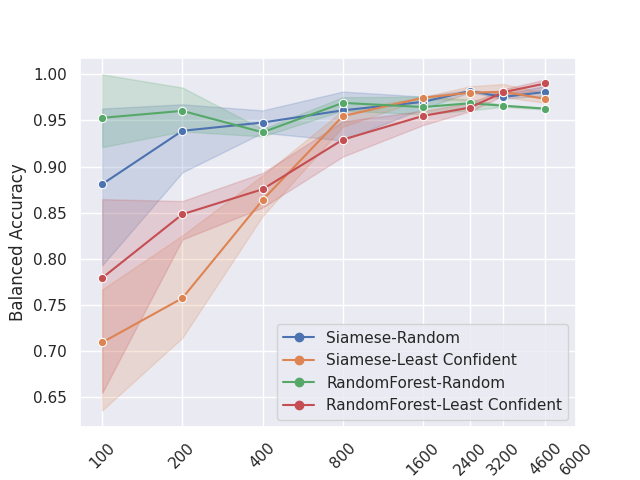}
        \caption{Active Learning: pre-train batch test.}
        \label{fig:pre_batch}   
    \end{subfigure}%
    \begin{subfigure}{.45\textwidth}
        \includegraphics[width=\linewidth]{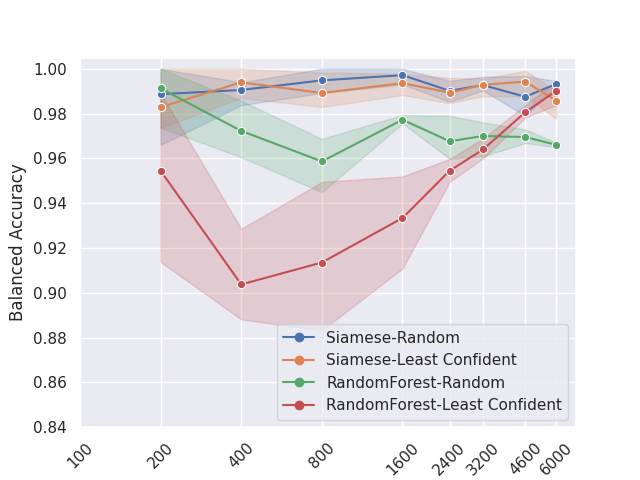}
        \caption{Active Learning: post-train batch test.} 
        \label{fig:post_batch}   
    \end{subfigure}

    \begin{subfigure}{.45\textwidth}
        \includegraphics[width=\linewidth]{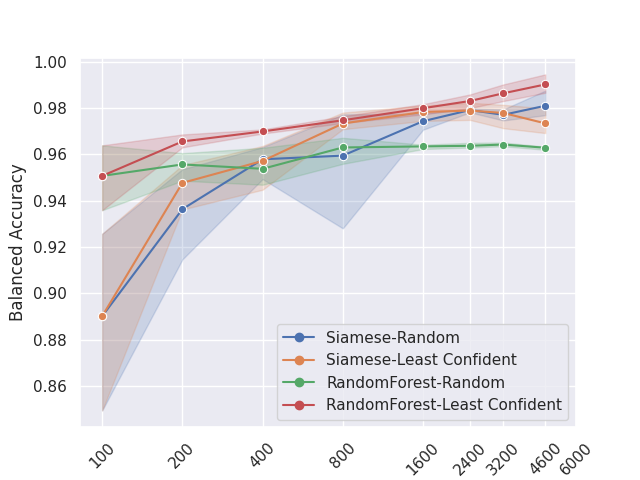}
        \caption{Active Learning: not-yet-labelled test.}
        \label{fig:unl}   
    \end{subfigure}%
    \begin{subfigure}{.45\textwidth}
        \includegraphics[width=\linewidth]{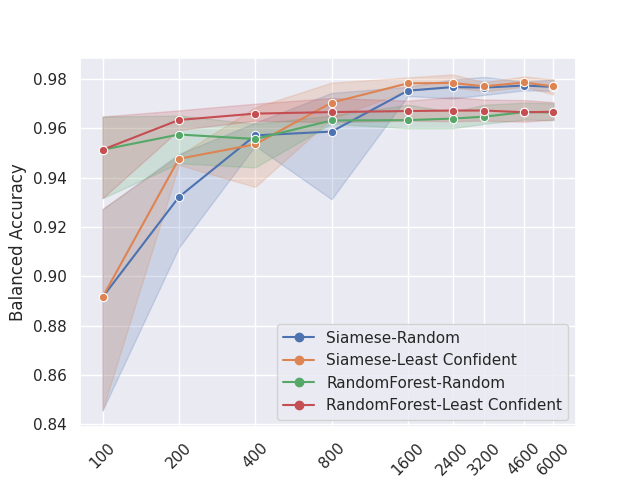}
        \caption{Active Learning: randomly ordered test.} 
        \label{fig:test}   
    \end{subfigure}        
    \caption{\textbf{Active Learning performances.} BA during the Active Learning procedure computed over: (\subref{fig:pre_batch}) next-to-be-labelled instances ($X^{j-1}_c$), (\subref{fig:post_batch}) just-labelled instances ($X^{j-1}_c$), (\subref{fig:unl}) all the remaining unlabelled instances ($X^{j-1}_u$), (\subref{fig:test}) the test set $S_\text{test}^\text{RO}$, as described in Section~\ref{sec:al_setting}. On the $x$-axis we plot the number of labelled samples (log scale), starting from $\lvert X^0_l\rvert=100$ up to $\lvert X^{M-2}_l\rvert=4600$ --- i.e. before adding the remaining $B^{M-1}$ samples in the last iteration. In the post-train case (\subref{fig:post_batch}), the evaluation starts from $\lvert X^1_l\rvert=200$ (i.e. after adding the first $B^{1}$ samples) up to the whole training dataset $\lvert X^{M-1}_l\rvert=6000$. The evaluation over $S_\text{test}^\text{RO}$ (\subref{fig:test}) is done in all available steps, i.e. from 100 up to 6000 samples. BA of Random Forest and Siamese Network models both as Random learners and as Least Confident learners is reported. The mean and 95\% normal confidence intervals are obtained by aggregating the BA over the 3 cross-validation folds.}
\end{figure}

As mentioned in Section~\ref{sec:al}, the choice of the subset of unlabelled instances to be labelled ($X_{c}^j$) lies at the heart of the Active Learning strategy. 
To benchmark this procedure, besides the \emph{Least Confident learner}~(LC) selecting $X_{c}^j$ according to the uncertainty (Equation~\eqref{eq:unc_noise}), we run the same experiment with a \emph{Random learner}~(R) picking the unlabelled samples in a purely random fashion. 
Therefore, we end up with a total of four learners. 
At the end of each iteration $j$, we evaluate their performance as follows: 
\begin{enumerate} 
    \item \emph{pre-train batch test}: we test the model $\mathcal{C}^{j}$ on the next-to-be-labelled instances, i.e. $X^{j}_{c}$: we here expect poor results for the LC learners, since we are testing on most uncertain samples for the model $\mathcal{C}^{j}$~(see Figure \ref{fig:pre_batch}).
    \item We train with respect to the new training set $(X^{j}_l , y^{j}_l)$, thus obtaining $\mathcal{C}^{j+1}$.
    \item We test $\mathcal{C}^{j+1}$ on:
    \begin{itemize}
        \item the batch samples $X^{j}_{c}$, and we refer to it as the \emph{post-train batch test} (see Figure~\ref{fig:post_batch}): we here expect good results, since it is an in-sample valuation;
        \item the updated unlabelled set $X^{j}_u$, i.e. all the remaining unlabelled instances, and we refer to it as the \emph{not-labelled-yet test} (see Figure~\ref{fig:unl});
        \item the actual test set, namely $S_{\text{test}}^\text{RO}$ (see Figure~\ref{fig:test}).
    \end{itemize}
\end{enumerate}


\section{Discussion}
\label{sec:discussion}


Table~\ref{tab:results} summarises the results of the experiments in terms of BA. The following is a list of insights we can derive from its inspection.

The Random Forest model trained on the small dataset $S^\text{S}_\text{train}$ has a good performance on the randomly ordered test set $S_\text{test}^\text{RO}$. This is likely due to the fact that the input features --- described in Section~\ref{sec:RF} --- are essentially string similarity measures, thus the pattern to be learned does not need a lot of observations. The downside is poor generalisation when more data are provided. Indeed, by increasing the training set size from $S^\text{S}_\text{train}$ to $S^\text{M}_\text{train}$, less than 2\% of BA is gained on the randomly ordered test set and no gain at all from $S^\text{M}_\text{train}$ to $S^\text{L}_\text{train}$. Moreover, its performance drops by $\sim 27\%$ and $\sim 24\%$ in the JW~ordered test and the domain shifted test, respectively.

The same thing is true --- even more so --- for the JW~based classifier. In this case, the decision tree needs only to perform an optimal choice of the threshold to put on the JW~similarity score. We can expect that the increase in size of the training set is only slightly changing this threshold, with a negligible impact on the out-of-sample performances.
Then we can deduce that JW metric does not tend to overfit his domain and can generalise with acceptable performances.
The same reasoning can be applied to the other string match models. In particular, the InDel ratio and the Token Set Ratio perform remarkably well with a small amount of data, again due to the simple rule to be learned by the classifier. Concerning the JW model, we can observe a poorer generalisation to new domains.

It is worth noticing that the drop in performance for the Baseline classifiers when switching from $S_\text{test}^\text{RO}$ to $S_\text{test}^\text{JO}$ and $S_\text{test}^\text{DS}$ is slightly larger than for the Random Forest. This is reasonable given that the Random Forest may use the information coming from all of the similarity metric scores at the same time.

The Siamese model systematically outperforms other approaches on both medium and large training datasets. This demonstrates the ability of the Neural Network to avoid overfitting to the specific domain, to generalise across different distributions, and to learn an alias associated with a company that may differ significantly from a simple string match similarity (e.g. the pair ``REF SRL'' and ``RENOVARE ENERGY FARM SRL''). 

Table~\ref{tab:cool_examples} shows several examples of matching and non-matching couples of company string names as they are classified by the Siamese model trained on $S_{\text{train}}^{\text{L}}$, with the corresponding estimated probability $\hat{y}$. 
In particular, the examples in Table~\ref{tab:cool_examples} are extracted by drawing from non-matching couples with \emph{high} JW~similarity, and from matching couples with \emph{low} JW~similarity. In this way, we expect to highlight some interesting and challenging sample couples.
Indeed, one can see that the Siamese model is able to correctly classify company names expressed as acronyms (e.g. in ``S.P.I.G.A.'' and ``REF''). On the other hand, there are cases more difficult to explain, such as the correctly classified match for the couple (``RONDA'', ``TORO ASSICURAZIONI SPA'') where the entity is indeed the same --- TORO insurance actually merged into RONDA in 2004\footnote{\href{https://www.gazzettaufficiale.it/atto/parte_seconda/caricaDettaglioAtto/originario?atto.dataPubblicazioneGazzetta=2004-04-08&atto.codiceRedazionale=S-7119}{gazzettaufficiale (GU Parte Seconda n.83 del 8-4-2004)}.} --- but the company names are completely different. Further work is needed to explain the reasons behind such counter-intuitive matches of the Neural Network, and to find the patterns behind such classifications.

We reported false positive examples where the names are very similar but they actually belong to different companies, e.g. ``E.U.R.O. S.R.L.'' and ``EURO STEEL SRL/MILANO''. The overconfidence of the prediction could be solved, e.g. by incorporating additional information, such as address, holding and legal form of the two companies.

As expected, most of the false negative samples --- i.e. pairs representing the same entity but predicted to be non-matching --- in Table~\ref{tab:cool_examples} can be related to situations in which the company names are (almost) completely different, but the entity is indeed the same, likely due to some legal transaction (merger, acquisition, consolidation, etc.) as discussed in Section~\ref{sec:data_labelling}.

Finally, the true negative examples show the remarkable capabilities of the Siamese model to correctly classify as different entities even pairs with very similar company names, such as ``RECOS S.R.L.'' and ``PECOS SRL'', with very high confidence.


Figure~\ref{fig:test} shows the out-of-sample BA of the Siamese Network and the Random Forest when computed on $S_\text{test}^\text{RO}$ at each Active Learning step. We can easily see that the Least Confident approach systematically outperforms the random choice for both models, confirming the value of Active Learning. However, while the Random Forest performance plateaus already before 400 samples, the Siamese needs up to 2,000: the Siamese Network has to leverage a larger amount of data to effectively learn patterns.

Figures~\ref{fig:pre_batch}-\ref{fig:post_batch} can help us understand the mechanisms at play. Figure~\ref{fig:pre_batch} displays out-of-sample BA values on next-to-be-labelled batches: we expect poor performance for the Least Confident choice with respect to Random choice since in the former case we are selecting uncertain instances (i.e. difficult for the model) on purpose. 
Figure~\ref{fig:post_batch}, on the other hand, displays BA values on batches just fed to the models: we here expect --- in general --- higher performances, being an in-sample evaluation. Interestingly, we see that the Least Confident choice has poor performance with respect to Random choice. This may be due to the fact that (at least a fraction of the) most uncertain observations remain indeed intrinsically difficult to classify, despite the training. This effect is more pronounced for the Random Forest, likely because it is largely based on similarity metrics. Notice that, as the number of samples seen by the model increases, this effect is less and less pronounced and finally reverts, indicating that residual samples are becoming easier to classify in the Least Confident approach.

Incidentally, we notice that comparing the performances of the model on a batch before and after the model has been trained on it, allows to define an \emph{early stopping} rule. Namely, calling $Acc_{pre}$ and $Acc_{post}$ the BA of a model over a batch before and after it has been trained on it, respectively, we can define a threshold $\theta$ and interrupt the process when: $ \lvert Acc_{post} - Acc_{pre} \rvert < \theta$.
The rationale is that,  if a model has about the same performances on a batch before and after having been trained on it, it means that it has already learned to generalise over unseen data. This rule can be applied, with an appropriate~$\theta$, both on the Least Confident and Random algorithm.

Finally, Figure~\ref{fig:unl} shows the out-of-sample BA when computed on $X_u$ (i.e. all the unlabelled samples of the training set at that step) at each step of Algorithm~\ref{alg:activelearning}. This plot confirms that the Least Confident approach chooses the instances in such a way that the remaining ones are simpler to be classified. This is especially true for Random~Forest.

\section{Conclusions}
\label{sec:conclusion}

In our analysis, we have compared the performances of several supervised classifiers in the field of company name disambiguation.
Providing pairs of company names as input, we consider two types of Machine Learning classifiers: Decision Stumps and Random Forest classifiers based on classical string similarity measures between the two names; a Neural Network classifier on top of a learned LSTM embedding space of strings.

The data are extracted from external company registry data and bank transfers. More specifically, we collect 9,000 couples of company names from external company registry data, and 200 couples of beneficiary names in bank transfers (the latter used only for testing purposes). 
All approaches are evaluated over three different test sets: a ``randomly ordered'' test set (RO), i.e. 3000 samples randomly chosen from the company registry dataset, a ``Jaro-Winkler ordered'' test set (JO), i.e. 200 instances drawn from RO in such a way that we select JW-dissimilar actual matches and JW-similar actual non-matches, and a ``domain shift" test set (DS), i.e. 200 couples of beneficiary names taken from a (different) dataset of bank transfers. 

The purpose of this work is twofold: on the one hand, we show that if enough data is available, the Siamese approach outperforms the other models and can be applied to other domains. Indeed, according to Table~\ref{tab:results}, the performance of the Baseline methods and the Random Forest barely improves when more data are provided for training. This is likely due to the fact that all the information extracted from string pairs is encoded in classical string similarity metrics for Baseline Trees and Random Forest. 

On the contrary, increasing the size of the training set improves the performance of the Siamese model, as it can learn a more effective embedding space the more data it learns from. This demonstrates the Neural Network's capacity to generalise while avoiding overfitting to a specific domain. These features enable our Siamese Neural Network to learn its own concept of string similarity and appropriately detect aliases connected with a company that differ significantly from a basic string match similarity.

The other goal of our research is to demonstrate how to make human labelling more efficient by using an Active Learning strategy.
Indeed, starting with a minimal training set of 100 labelled data, we show that training the model with subsequent batches of the most uncertain samples (Least Confident learner) is more efficient than training with randomly chosen instances. 

One limitation of this work is the use of company names only for the goal of disambiguation. As previously said, it is possible that company names by themselves \emph{do not} contain enough information to resolve all the matches, as in cases where completely different company names still refer to the same Entity. We leave to future work the extension to include additional information, such as addresses, legal forms, shareholding, etc. We also plan to perform a more thorough analysis of the architecture of the Siamese Recurrent Network, possibly using a Transformer approach for the input sequences~\citep{vaswani2017attention}. Given the effectiveness of the Active Learning procedure, we plan to use an unlabelled dataset of hundreds of thousands of Entity pairs to extract from them the most informative few thousand pairs to label, to analyse how the procedure here outlined scales with data availability.

\section*{Author contributions}
\textbf{\textsl{Riccardo~Crupi}}: Conceptualisation, Methodology, Software, Validation, Investigation, Writing - Original Draft, Project Administration. 
\textbf{\textsl{Michele~Grasso}}: Software, Validation, Investigation, Writing - Original Draft. 
\textbf{\textsl{Daniele~Regoli}}: Methodology, Investigation, Writing - Original Draft. 
\textbf{\textsl{Shuyi~Yang}}: Methodology, Software, Investigation, Writing - Review \& Editing. 
\textbf{\textsl{Simone~Scarsi}}: Methodology, Software, Investigation, Writing - Review \& Editing. 
\textbf{\textsl{Alessandro~Mercanti}}: Data Curation, Writing - Review \& Editing.
\textbf{\textsl{Alessandro~Basile}}: Writing - Review \& Editing.
\textbf{\textsl{Andrea~Cosentini}}: Writing - Review \& Editing, Supervision. 

\section*{Acknowledgements}
We would like to thank Ilaria Penco for her assistance with the legal aspects of the manuscript. We thank Andrea Barral for useful discussions on methodology and data curation. We also thank Giacomo Di Prinzio, Giulia Genta and  Nives Visentin from Intesa Sanpaolo, and Sandro Bellu, Indrit Gjonaj, Andrea Giordano and Gabriele Pellegrinetti from Tecnet Dati s.r.l., namely the team that developed the disambiguation project for Intesa Sanpaolo, that inspired the research here presented.

\bibliography{references.bib}

\begin{thebibliography}{30}
\expandafter\ifx\csname natexlab\endcsname\relax\def\natexlab#1{#1}\fi
\providecommand{\url}[1]{\texttt{#1}}
\providecommand{\href}[2]{#2}
\providecommand{\path}[1]{#1}
\providecommand{\DOIprefix}{doi:}
\providecommand{\ArXivprefix}{arXiv:}
\providecommand{\URLprefix}{URL: }
\providecommand{\Pubmedprefix}{pmid:}
\providecommand{\doi}[1]{\href{http://dx.doi.org/#1}{\path{#1}}}
\providecommand{\Pubmed}[1]{\href{pmid:#1}{\path{#1}}}
\providecommand{\bibinfo}[2]{#2}
\ifx\xfnm\relax \def\xfnm[#1]{\unskip,\space#1}\fi
\bibitem[{Abadi et~al.(2015)Abadi, Agarwal, Barham, Brevdo, Chen, Citro,
  Corrado, Davis, Dean, Devin, Ghemawat, Goodfellow, Harp, Irving, Isard, Jia,
  Jozefowicz, Kaiser, Kudlur, Levenberg, Man\'{e}, Monga, Moore, Murray, Olah,
  Schuster, Shlens, Steiner, Sutskever, Talwar, Tucker, Vanhoucke, Vasudevan,
  Vi\'{e}gas, Vinyals, Warden, Wattenberg, Wicke, Yu and
  Zheng}]{tensorflow2015-whitepaper}
\bibinfo{author}{Abadi, M.}, \bibinfo{author}{Agarwal, A.},
  \bibinfo{author}{Barham, P.}, \bibinfo{author}{Brevdo, E.},
  \bibinfo{author}{Chen, Z.}, \bibinfo{author}{Citro, C.},
  \bibinfo{author}{Corrado, G.S.}, \bibinfo{author}{Davis, A.},
  \bibinfo{author}{Dean, J.}, \bibinfo{author}{Devin, M.},
  \bibinfo{author}{Ghemawat, S.}, \bibinfo{author}{Goodfellow, I.},
  \bibinfo{author}{Harp, A.}, \bibinfo{author}{Irving, G.},
  \bibinfo{author}{Isard, M.}, \bibinfo{author}{Jia, Y.},
  \bibinfo{author}{Jozefowicz, R.}, \bibinfo{author}{Kaiser, L.},
  \bibinfo{author}{Kudlur, M.}, \bibinfo{author}{Levenberg, J.},
  \bibinfo{author}{Man\'{e}, D.}, \bibinfo{author}{Monga, R.},
  \bibinfo{author}{Moore, S.}, \bibinfo{author}{Murray, D.},
  \bibinfo{author}{Olah, C.}, \bibinfo{author}{Schuster, M.},
  \bibinfo{author}{Shlens, J.}, \bibinfo{author}{Steiner, B.},
  \bibinfo{author}{Sutskever, I.}, \bibinfo{author}{Talwar, K.},
  \bibinfo{author}{Tucker, P.}, \bibinfo{author}{Vanhoucke, V.},
  \bibinfo{author}{Vasudevan, V.}, \bibinfo{author}{Vi\'{e}gas, F.},
  \bibinfo{author}{Vinyals, O.}, \bibinfo{author}{Warden, P.},
  \bibinfo{author}{Wattenberg, M.}, \bibinfo{author}{Wicke, M.},
  \bibinfo{author}{Yu, Y.}, \bibinfo{author}{Zheng, X.}, \bibinfo{year}{2015}.
\newblock \bibinfo{title}{{TensorFlow}: Large-scale machine learning on
  heterogeneous systems}.
\newblock \URLprefix \url{https://www.tensorflow.org/}. \bibinfo{note}{software
  available from tensorflow.org}.
\bibitem[{Aggarwal et~al.(2014)Aggarwal, Kong, Gu, Han and
  Philip}]{aggarwal2014active}
\bibinfo{author}{Aggarwal, C.C.}, \bibinfo{author}{Kong, X.},
  \bibinfo{author}{Gu, Q.}, \bibinfo{author}{Han, J.}, \bibinfo{author}{Philip,
  S.Y.}, \bibinfo{year}{2014}.
\newblock \bibinfo{title}{Active learning: A survey}, in:
  \bibinfo{booktitle}{Data Classification}. \bibinfo{publisher}{Chapman and
  Hall/CRC}, pp. \bibinfo{pages}{599--634}.
\bibitem[{Aghaebrahimian and Cieliebak(2020)}]{aghaebrahimian2020named}
\bibinfo{author}{Aghaebrahimian, A.}, \bibinfo{author}{Cieliebak, M.},
  \bibinfo{year}{2020}.
\newblock \bibinfo{title}{Named entity disambiguation at scale}, in:
  \bibinfo{booktitle}{IAPR Workshop on Artificial Neural Networks in Pattern
  Recognition}, \bibinfo{organization}{Springer}. pp.
  \bibinfo{pages}{102--110}.
\bibitem[{Arora and Agarwal(2007)}]{arora2007active}
\bibinfo{author}{Arora, S.}, \bibinfo{author}{Agarwal, S.},
  \bibinfo{year}{2007}.
\newblock \bibinfo{title}{Active learning for natural language processing}.
\newblock \bibinfo{journal}{Language Technologies Institute School of Computer
  Science Carnegie Mellon University} .
\bibitem[{Banko and Brill(2001)}]{banko2001scaling}
\bibinfo{author}{Banko, M.}, \bibinfo{author}{Brill, E.}, \bibinfo{year}{2001}.
\newblock \bibinfo{title}{Scaling to very very large corpora for natural
  language disambiguation}, in: \bibinfo{booktitle}{Proceedings of the 39th
  annual meeting of the Association for Computational Linguistics}, pp.
  \bibinfo{pages}{26--33}.
\bibitem[{Barlaug and Gulla(2021)}]{barlaug2021neural}
\bibinfo{author}{Barlaug, N.}, \bibinfo{author}{Gulla, J.A.},
  \bibinfo{year}{2021}.
\newblock \bibinfo{title}{Neural networks for entity matching: A survey}.
\newblock \bibinfo{journal}{ACM Transactions on Knowledge Discovery from Data
  (TKDD)} \bibinfo{volume}{15}, \bibinfo{pages}{1--37}.
\bibitem[{Bromley et~al.(1993)Bromley, Guyon, LeCun, S{\"a}ckinger and
  Shah}]{bromley1993signature}
\bibinfo{author}{Bromley, J.}, \bibinfo{author}{Guyon, I.},
  \bibinfo{author}{LeCun, Y.}, \bibinfo{author}{S{\"a}ckinger, E.},
  \bibinfo{author}{Shah, R.}, \bibinfo{year}{1993}.
\newblock \bibinfo{title}{Signature verification using a ``siamese'' time delay
  neural network}.
\newblock \bibinfo{journal}{Advances in neural information processing systems}
  \bibinfo{volume}{6}.
\bibitem[{Chicco(2021)}]{chicco2021siamese}
\bibinfo{author}{Chicco, D.}, \bibinfo{year}{2021}.
\newblock \bibinfo{title}{Siamese neural networks: An overview}.
\newblock \bibinfo{journal}{Artificial neural networks} ,
  \bibinfo{pages}{73--94}.
\bibitem[{Chicco et~al.(2021)Chicco, T{\"o}tsch and
  Jurman}]{chicco2021matthews}
\bibinfo{author}{Chicco, D.}, \bibinfo{author}{T{\"o}tsch, N.},
  \bibinfo{author}{Jurman, G.}, \bibinfo{year}{2021}.
\newblock \bibinfo{title}{{The Matthews correlation coefficient (MCC) is more
  reliable than balanced accuracy, bookmaker informedness, and markedness in
  two-class confusion matrix evaluation}}.
\newblock \bibinfo{journal}{BioData mining} \bibinfo{volume}{14},
  \bibinfo{pages}{1--22}.
\bibitem[{Christen(2006)}]{christen2006comparison}
\bibinfo{author}{Christen, P.}, \bibinfo{year}{2006}.
\newblock \bibinfo{title}{A comparison of personal name matching: Techniques
  and practical issues}, in: \bibinfo{booktitle}{Sixth IEEE International
  Conference on Data Mining-Workshops (ICDMW'06)},
  \bibinfo{organization}{IEEE}. pp. \bibinfo{pages}{290--294}.
\bibitem[{Cohen et~al.(2003)Cohen, Ravikumar, Fienberg
  et~al.}]{cohen2003comparison}
\bibinfo{author}{Cohen, W.W.}, \bibinfo{author}{Ravikumar, P.},
  \bibinfo{author}{Fienberg, S.E.}, et~al., \bibinfo{year}{2003}.
\newblock \bibinfo{title}{A comparison of string distance metrics for
  name-matching tasks.}, in: \bibinfo{booktitle}{IIWeb}, pp.
  \bibinfo{pages}{73--78}.
\bibitem[{van Engelen and Hoos(2020)}]{DBLP:journals/ml/EngelenH20}
\bibinfo{author}{van Engelen, J.E.}, \bibinfo{author}{Hoos, H.H.},
  \bibinfo{year}{2020}.
\newblock \bibinfo{title}{A survey on semi-supervised learning}.
\newblock \bibinfo{journal}{Mach. Learn.} \bibinfo{volume}{109},
  \bibinfo{pages}{373--440}.
\newblock \URLprefix \url{https://doi.org/10.1007/s10994-019-05855-6},
  \DOIprefix\doi{10.1007/s10994-019-05855-6}.
\bibitem[{Hastie et~al.(2009)Hastie, Tibshirani, Friedman and
  Friedman}]{hastie2009elements}
\bibinfo{author}{Hastie, T.}, \bibinfo{author}{Tibshirani, R.},
  \bibinfo{author}{Friedman, J.H.}, \bibinfo{author}{Friedman, J.H.},
  \bibinfo{year}{2009}.
\newblock \bibinfo{title}{The elements of statistical learning: data mining,
  inference, and prediction}. volume~\bibinfo{volume}{2}.
\newblock \bibinfo{publisher}{Springer}.
\bibitem[{Huang(2021)}]{huang2021deepal}
\bibinfo{author}{Huang, K.H.}, \bibinfo{year}{2021}.
\newblock \bibinfo{title}{Deepal: Deep active learning in python}.
\newblock \bibinfo{journal}{arXiv preprint arXiv:2111.15258} .
\bibitem[{Kolitsas et~al.(2018)Kolitsas, Ganea and Hofmann}]{kolitsas2018end}
\bibinfo{author}{Kolitsas, N.}, \bibinfo{author}{Ganea, O.E.},
  \bibinfo{author}{Hofmann, T.}, \bibinfo{year}{2018}.
\newblock \bibinfo{title}{End-to-end neural entity linking}.
\newblock \bibinfo{journal}{arXiv preprint arXiv:1808.07699} .
\bibitem[{Meduri et~al.(2020)Meduri, Popa, Sen and
  Sarwat}]{meduri2020comprehensive}
\bibinfo{author}{Meduri, V.V.}, \bibinfo{author}{Popa, L.},
  \bibinfo{author}{Sen, P.}, \bibinfo{author}{Sarwat, M.},
  \bibinfo{year}{2020}.
\newblock \bibinfo{title}{A comprehensive benchmark framework for active
  learning methods in entity matching}, in: \bibinfo{booktitle}{Proceedings of
  the 2020 ACM SIGMOD International Conference on Management of Data}, pp.
  \bibinfo{pages}{1133--1147}.
\bibitem[{Navarro(2001)}]{navarro2001guided}
\bibinfo{author}{Navarro, G.}, \bibinfo{year}{2001}.
\newblock \bibinfo{title}{A guided tour to approximate string matching}.
\newblock \bibinfo{journal}{ACM computing surveys (CSUR)} \bibinfo{volume}{33},
  \bibinfo{pages}{31--88}.
\bibitem[{Neculoiu et~al.(2016)Neculoiu, Versteegh and
  Rotaru}]{neculoiu2016learning}
\bibinfo{author}{Neculoiu, P.}, \bibinfo{author}{Versteegh, M.},
  \bibinfo{author}{Rotaru, M.}, \bibinfo{year}{2016}.
\newblock \bibinfo{title}{Learning text similarity with siamese recurrent
  networks}, in: \bibinfo{booktitle}{Proceedings of the 1st Workshop on
  Representation Learning for NLP}, pp. \bibinfo{pages}{148--157}.
\bibitem[{Pedregosa et~al.(2011)Pedregosa, Varoquaux, Gramfort, Michel,
  Thirion, Grisel, Blondel, Prettenhofer, Weiss, Dubourg, Vanderplas, Passos,
  Cournapeau, Brucher, Perrot and Duchesnay}]{scikit-learn}
\bibinfo{author}{Pedregosa, F.}, \bibinfo{author}{Varoquaux, G.},
  \bibinfo{author}{Gramfort, A.}, \bibinfo{author}{Michel, V.},
  \bibinfo{author}{Thirion, B.}, \bibinfo{author}{Grisel, O.},
  \bibinfo{author}{Blondel, M.}, \bibinfo{author}{Prettenhofer, P.},
  \bibinfo{author}{Weiss, R.}, \bibinfo{author}{Dubourg, V.},
  \bibinfo{author}{Vanderplas, J.}, \bibinfo{author}{Passos, A.},
  \bibinfo{author}{Cournapeau, D.}, \bibinfo{author}{Brucher, M.},
  \bibinfo{author}{Perrot, M.}, \bibinfo{author}{Duchesnay, E.},
  \bibinfo{year}{2011}.
\newblock \bibinfo{title}{Scikit-learn: Machine learning in {P}ython}.
\newblock \bibinfo{journal}{Journal of Machine Learning Research}
  \bibinfo{volume}{12}, \bibinfo{pages}{2825--2830}.
\bibitem[{Piskorski and Jacquet(2020)}]{piskorski2020tf}
\bibinfo{author}{Piskorski, J.}, \bibinfo{author}{Jacquet, G.},
  \bibinfo{year}{2020}.
\newblock \bibinfo{title}{{TF-IDF Character N-grams} versus word
  embedding-based models for fine-grained event classification: a preliminary
  study}, in: \bibinfo{booktitle}{Proceedings of the Workshop on Automated
  Extraction of Socio-political Events from News 2020}, pp.
  \bibinfo{pages}{26--34}.
\bibitem[{Prasetya et~al.(2018)Prasetya, Wibawa and
  Hirashima}]{prasetya2018performance}
\bibinfo{author}{Prasetya, D.D.}, \bibinfo{author}{Wibawa, A.P.},
  \bibinfo{author}{Hirashima, T.}, \bibinfo{year}{2018}.
\newblock \bibinfo{title}{The performance of text similarity algorithms}.
\newblock \bibinfo{journal}{International Journal of Advances in Intelligent
  Informatics} \bibinfo{volume}{4}, \bibinfo{pages}{63--69}.
\bibitem[{Ren et~al.(2022)Ren, Xiao, Chang, Huang, Li, Gupta, Chen and
  Wang}]{DBLP:journals/csur/RenXCHLGCW22}
\bibinfo{author}{Ren, P.}, \bibinfo{author}{Xiao, Y.}, \bibinfo{author}{Chang,
  X.}, \bibinfo{author}{Huang, P.}, \bibinfo{author}{Li, Z.},
  \bibinfo{author}{Gupta, B.B.}, \bibinfo{author}{Chen, X.},
  \bibinfo{author}{Wang, X.}, \bibinfo{year}{2022}.
\newblock \bibinfo{title}{A survey of deep active learning}.
\newblock \bibinfo{journal}{{ACM} Comput. Surv.} \bibinfo{volume}{54},
  \bibinfo{pages}{180:1--180:40}.
\newblock \URLprefix \url{https://doi.org/10.1145/3472291},
  \DOIprefix\doi{10.1145/3472291}.
\bibitem[{Santos et~al.(2018)Santos, Murrieta-Flores, Calado and
  Martins}]{santos2018toponym}
\bibinfo{author}{Santos, R.}, \bibinfo{author}{Murrieta-Flores, P.},
  \bibinfo{author}{Calado, P.}, \bibinfo{author}{Martins, B.},
  \bibinfo{year}{2018}.
\newblock \bibinfo{title}{Toponym matching through deep neural networks}.
\newblock \bibinfo{journal}{International Journal of Geographical Information
  Science} \bibinfo{volume}{32}, \bibinfo{pages}{324--348}.
\bibitem[{Sen et~al.(2020)Sen, Hajra and Ghosh}]{sen2020supervised}
\bibinfo{author}{Sen, P.C.}, \bibinfo{author}{Hajra, M.},
  \bibinfo{author}{Ghosh, M.}, \bibinfo{year}{2020}.
\newblock \bibinfo{title}{Supervised classification algorithms in machine
  learning: A survey and review}, in: \bibinfo{booktitle}{Emerging technology
  in modelling and graphics}. \bibinfo{publisher}{Springer}, pp.
  \bibinfo{pages}{99--111}.
\bibitem[{Settles(2009)}]{settles2009active}
\bibinfo{author}{Settles, B.}, \bibinfo{year}{2009}.
\newblock \bibinfo{title}{Active learning literature survey} .
\bibitem[{Sorscher et~al.(2022)Sorscher, Geirhos, Shekhar, Ganguli and
  Morcos}]{sorscher2022beyond}
\bibinfo{author}{Sorscher, B.}, \bibinfo{author}{Geirhos, R.},
  \bibinfo{author}{Shekhar, S.}, \bibinfo{author}{Ganguli, S.},
  \bibinfo{author}{Morcos, A.S.}, \bibinfo{year}{2022}.
\newblock \bibinfo{title}{Beyond neural scaling laws: beating power law scaling
  via data pruning}.
\newblock \bibinfo{journal}{arXiv preprint arXiv:2206.14486} .
\bibitem[{Sun et~al.(2015)Sun, Ma and Wang}]{sun2015comparative}
\bibinfo{author}{Sun, Y.}, \bibinfo{author}{Ma, L.}, \bibinfo{author}{Wang,
  S.}, \bibinfo{year}{2015}.
\newblock \bibinfo{title}{A comparative evaluation of string similarity metrics
  for ontology alignment}.
\newblock \bibinfo{journal}{Journal of Information \&Computational Science}
  \bibinfo{volume}{12}, \bibinfo{pages}{957--964}.
\bibitem[{Vaswani et~al.(2017)Vaswani, Shazeer, Parmar, Uszkoreit, Jones,
  Gomez, Kaiser and Polosukhin}]{vaswani2017attention}
\bibinfo{author}{Vaswani, A.}, \bibinfo{author}{Shazeer, N.},
  \bibinfo{author}{Parmar, N.}, \bibinfo{author}{Uszkoreit, J.},
  \bibinfo{author}{Jones, L.}, \bibinfo{author}{Gomez, A.N.},
  \bibinfo{author}{Kaiser, {\L}.}, \bibinfo{author}{Polosukhin, I.},
  \bibinfo{year}{2017}.
\newblock \bibinfo{title}{Attention is all you need}.
\newblock \bibinfo{journal}{Advances in neural information processing systems}
  \bibinfo{volume}{30}.
\bibitem[{Zhan et~al.(2022)Zhan, Wang, Huang, Xiong, Dou and
  Chan}]{zhan2022comparative}
\bibinfo{author}{Zhan, X.}, \bibinfo{author}{Wang, Q.}, \bibinfo{author}{Huang,
  K.h.}, \bibinfo{author}{Xiong, H.}, \bibinfo{author}{Dou, D.},
  \bibinfo{author}{Chan, A.B.}, \bibinfo{year}{2022}.
\newblock \bibinfo{title}{A comparative survey of deep active learning}.
\newblock \bibinfo{journal}{arXiv preprint arXiv:2203.13450} .
\bibitem[{Zhou(2018)}]{zhou2018brief}
\bibinfo{author}{Zhou, Z.H.}, \bibinfo{year}{2018}.
\newblock \bibinfo{title}{A brief introduction to weakly supervised learning}.
\newblock \bibinfo{journal}{National science review} \bibinfo{volume}{5},
  \bibinfo{pages}{44--53}.

\end{thebibliography}

\newgeometry{margin=1cm} 
\begin{landscape}
\thispagestyle{empty}
\begin{table}
\begin{center}
\ra{1.2}
\resizebox{0.8\linewidth}{!}{
\begin{tabular}{@{}l c c l c r c@{}}
\toprule
 && \multicolumn{2}{c}{\textbf{Predicted positives} ($\hat{y}\geq0.5$)} && \multicolumn{2}{c}{\textbf{Predicted negatives} ($\hat{y}<0.5$)} \\[10pt]
 
 & & $\hat{y}$ & company name pairs $\left(\begin{smallmatrix}a\\b\end{smallmatrix}\right)$ & \phantom{abc} & company name pairs $\left(\begin{smallmatrix}a\\b\end{smallmatrix}\right)$ & $\hat{y}$ \\
\cmidrule{3-7}
\multirow{12}{*}{\makecell[c]{\\[30pt]\textbf{same}\\ \textbf{entity}\\\textbf{($\bm{y=1}$)}}} & \phantom{abc} & \multirow{2}{*}{0.97}& S.P.I.G.A. S.R.L. && FOGLIATA S.P.A. & \multirow{2}{*}{0.006}\\
 &&& SPIGA-SOCIETA' PRODUZIONE E IMPORTAZIONE GENERALI ALIMENTARI-SRL && EDILCOS SRL & \\[10pt]

 && \multirow{2}{*}{0.69} & REF SRL && CREARE IN FOSSATO SRL IN LIQUIDAZIONE & \multirow{2}{*}{0.001}\\
 &&& RENOVARE ENERGY FARM SRL && WALD SRL & \\[10pt]
 
 && \multirow{2}{*}{0.94} & RONDA && UNIEURO S.P.A. & \multirow{2}{*}{0.002}\\
 &&& TORO ASSICURAZIONI SPA && SGM DISTRIBUZIONE SRL & \\[10pt]

 && \multirow{2}{*}{0.88} & SEAWAYS && MANUFATTI EPIS SRL & \multirow{2}{*}{0.02}\\
 &&& AGENZIA MARITTIMA LE NAVI-SEAWAYS SRL && EPIS SANTINO SNC DI EPIS STEFANO EC & \\[10pt]

 && \multirow{2}{*}{0.77} & TOSCOSERVICE LOGSTICA E SEVIZI SOCIETA' COOPERATIVA && DEA S.R.L. & \multirow{2}{*}{0.001}\\
 &&& IL GIRASOLE 2002-SC && UNENDO ENERGIA SUD SRL & \\[10pt]

 && \multirow{2}{*}{0.99} & ARGO S.R.L. && GIULIANO VINCENZA & \multirow{2}{*}{0.05}\\
 &&& ARGO SAS DI DI BIAGI FILIPPO && G\&P SRL & \\[10pt]

\cmidrule{3-7}

\multirow{12}{*}{\makecell[c]{\\[30pt]\textbf{different}\\ \textbf{entity}\\\textbf{($\bm{y=0}$)}}} && \multirow{2}{*}{0.99}& E.U.R.O. S.R.L. && EMMA SRL & \multirow{2}{*}{0.07}\\
 &&& EURO STEEL SRL/MILANO && GEMMA SRL & \\[10pt]

 && \multirow{2}{*}{0.91} & G.T.M. S.R.L. && SMV COSTRUZIONI SRL & \multirow{2}{*}{0.001}\\
 &&& GMC-SRL && RM COSTRUZIONI SRL & \\[10pt]
 
 && \multirow{2}{*}{0.56} & MARPER S.R.L. && RECOS S.R.L. & \multirow{2}{*}{0.002}\\
 &&& MAVIT SRL && PECOS SRL & \\[10pt]

 && \multirow{2}{*}{0.71} & MEA S.R.L. && TEXERA SRL & \multirow{2}{*}{0.2}\\
 &&& MAR - ALEO SRL && TER SRL & \\[10pt]

 && \multirow{2}{*}{0.88} & NUOVA REKORD S.R.L. && MEPRA SPA & \multirow{2}{*}{0.02}\\
 &&& NOVA LUX SRL/VERONA && PELMA SPA & \\[10pt]

 && \multirow{2}{*}{0.99} & FARMACIA WAGNER && ALIMA SRL & \multirow{2}{*}{0.08}\\
 &&& FARMACONSULT SRL && DALMA SRL & \\[10pt]

\bottomrule
\end{tabular}
}
\caption{\textbf{Representative examples}. Illustrative examples of results of the Siamese Neural Network trained on the large training set $S_\text{train}^L$. Predicted matches/non-matches are shown (together with the positive class predicted probability $\hat{y}$) for actual matching and non-matching pairs. The examples $\left(\begin{smallmatrix}a\\b\end{smallmatrix}\right)$ are selected from the \emph{randomly ordered} test set $S_\text{test}^\text{RO}$ (i.e. from the external company registry datasets). The specific model used to obtain these examples incorrectly classifies the $1.17\%$ of the positive cases and the $ 0.96\%$ of the negative cases tested.}
\label{tab:cool_examples}
\end{center}
\end{table}
\end{landscape}
\restoregeometry   

\appendix
\section{Ranfom~Forest feature importance}
\label{apx:feature_importance}
In this appendix, we analyse the importance of the features used by the Random~Forest classifier. Decision trees are used to divide data into relevant categories based on optimal splits. As a measure of how far the model deviates from a pure division, Gini impurity calculates the likelihood that a randomly selected example will be erroneously categorised by a certain node.
The Random Forest technique computes feature significance using the mean decrease impurity (MDI) or Gini importance, which is defined as the total node impurity weighted by the likelihood of reaching that node and averaged over all trees in the ensemble.

\begin{figure*}[ht]
     \centering
     \begin{subfigure}{.5\textwidth}
        \centering
         \includegraphics[width=\linewidth]{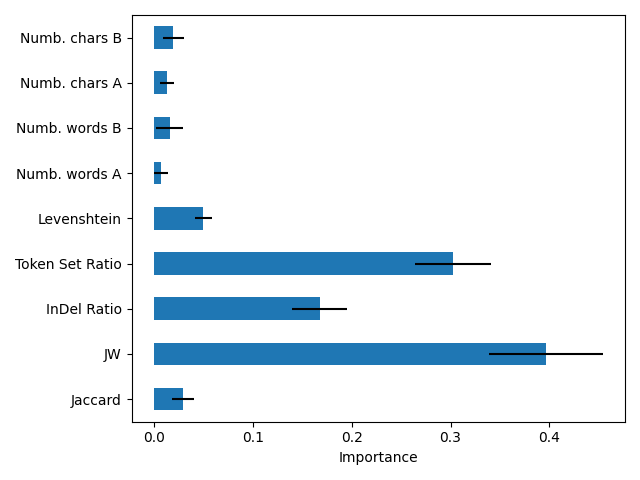}
         \caption{Random Forest trained on $S_\text{train}^\text{S}$}
         \label{fig:features_small}
     \end{subfigure}%
     \begin{subfigure}{0.5\textwidth}
        \centering
         \includegraphics[width=\textwidth]{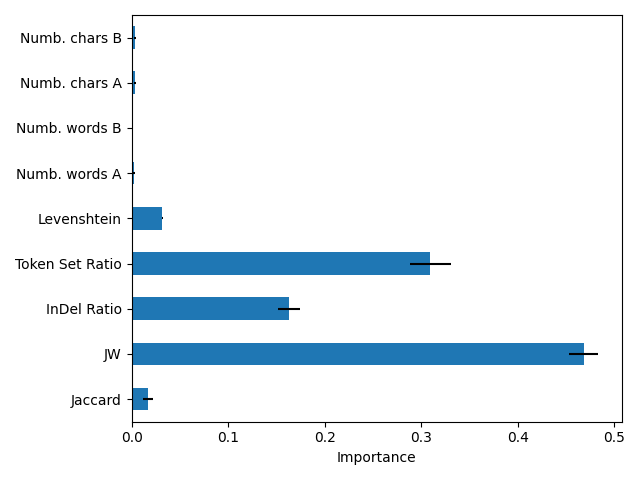}
         \caption{Random~Forest trained on $S_\text{train}^\text{M}$}
         \label{fig:features_medium}
     \end{subfigure}%
     \\
     \begin{subfigure}{0.5\textwidth}
         \centering
         \includegraphics[width=\textwidth]{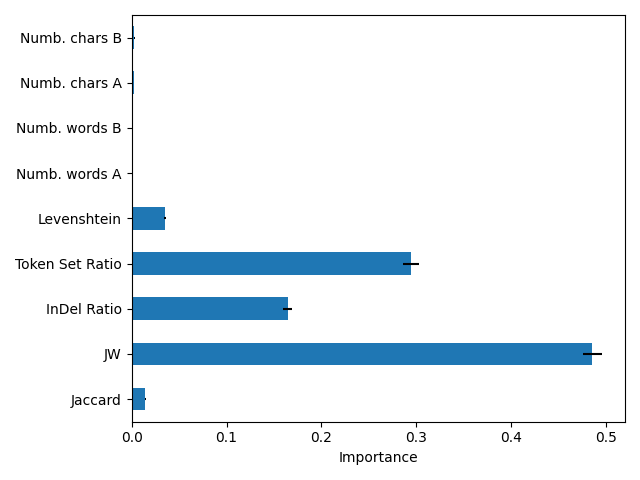}
         \caption{Random Forest trained on $S_\text{train}^\text{L}$}
         \label{fig:features_large}
     \end{subfigure}
        \caption{\textbf{Feature importance} for Random Forest classifiers trained on datasets of increasing size, namely small~(\subref{fig:features_small}), medium~(\subref{fig:features_medium}), and large~(\subref{fig:features_large}). Bars and corresponding error intervals are computed as the mean and standard deviation obtained over 3 cross-validation folds.}
        \label{fig:features_importance}
\end{figure*}

The bar plots in Figure~\ref{fig:features_importance} show the features importance of the Random~Forest model described in Sec. \ref{sec:RF} trained on the small (Figure~\ref{fig:features_small}), medium (Figure~\ref{fig:features_medium}), and large (Figure~\ref{fig:features_large}) training sets.
The significance of the features from the three cross-validation trials was aggregated and the resulting bars (in blue) and their error bars (in black) represent the mean and standard deviation, respectively. 

As can be seen from Figure~\ref{fig:features_importance}, the number of characters and the number of words of the two company names have negligible importance on the classifier. In particular, these features become less and less significant as more data are supplied for training.
JW~similarity is the most important feature used by the Random Forest classifier: the Random Forest trained on $S_\text{train}^\text{S}$ attributes $40\%$ of importance to JW~similarity, and this further increases as the classifier is trained on more data, reaching almost $50\%$ for the classifier trained on $S_\text{train}^\text{L}$.
The importance of the remaining features does not change much with the training set size, with a contribution of $\sim30\%$ for the Token Set Ratio, $\sim20\%$ for the InDel ratio, $\sim4\%$ for the Levensthein, and $\sim1\%$ for the Jaccard.

\section{Supervised classification performances}
\label{apx:other_metrics}
As discussed in Section~\ref{sec:experiments}, Table~\ref{tab:results} shows the out-of-sample BA on the test sets $S_{\text{test}}^\text{RO}$, $S_{\text{test}}^\text{JO}$, and $S_{\text{test}}^\text{DS}$. The BA ensures a correct evaluation of the classifiers' performance by weighting the accuracy of the classifier on each class by the number of observations in that class. This gives a more accurate picture of the classifier's performance in each class, even when the classes are imbalanced.

Along with BA, other metrics are commonly used to assess the quality of a model; here we will discuss $F_1$~score and MCC.
$F_1$~score is a very popular metric in the Machine Learning community since it combines precision and recall --- often two paramount aspects for model evaluation --- into a single metric. Indeed, it is defined as the weighted average of precision and recall.
MCC is instead a measure of the correlation between predicted and actual binary class labels and it ranges between~-1 --- indicating a total disagreement between prediction and observation --- and~1, indicating a perfect prediction.
While the $F_1$ score and MCC consider both precision and recall, BA is the average of the recall over all classes.
Therefore, also according to \cite{chicco2021matthews}, BA may be preferable when correctly classifying the ground truth instances (recall) is more important then making correct predictions (precision).

For completeness, in Table~\ref{tab:complete_results} we extend the results of Table~\ref{tab:results} by including the $F_1$~score and the MCC.

Results from these additional metrics largely confirm the ones discussed for BA, namely the fact that the Siamese Network approach systematically outperforms all other methods as long as enough data are provided for learning. This is especially true for the most difficult test sets, namely $S_\text{test}^\text{JO}$ and $S_\text{test}^\text{DS}$. 



\newgeometry{margin=1cm} 
\begin{landscape}
\thispagestyle{empty}
\begin{table}[p]
\begin{center}
\resizebox{\linewidth}{!}{%
\ra{1.3}
\begin{tabular}{@{}l c c c c c c c c c c c c c c c c c c c c c c c c c c c c c@{}}
\toprule
\makecell[l]{\textbf{training}\\ \textbf{set size}} & \makecell[c]{\textbf{test set}\\ \textbf{type}} & \phantom{ab} & \multicolumn{3}{c}{\textbf{Levenshtein}} & \phantom{ab} & \multicolumn{3}{c}{\textbf{InDel Ratio}} & \phantom{ab} & \multicolumn{3}{c}{\makecell[c]{\textbf{Token Set}\\ \textbf{Ratio}}} & \phantom{ab} & \multicolumn{3}{c}{\textbf{Jaccard}} & \phantom{ab} & \multicolumn{3}{c}{\textbf{JW}} & \phantom{ab} & \multicolumn{3}{c}{\makecell[c]{\textbf{Random}\\ \textbf{Forest}}} & \phantom{ab} & \multicolumn{3}{c}{\makecell[c]{\textbf{Siamese}\\ \textbf{Network}}}\\

\midrule

&&& BA & $F_1$ & MCC && BA & $F_1$ & MCC && BA & $F_1$ & MCC && BA & $F_1$ & MCC && BA & $F_1$ & MCC && BA & $F_1$ & MCC && BA & $F_1$ & MCC\\
\cmidrule{4-6}\cmidrule{8-10}\cmidrule{12-14}\cmidrule{16-18}\cmidrule{20-22}\cmidrule{24-26}\cmidrule{28-30}
 
\multirow{6}{*}{small} & \multirow{2}{*}{\textit{RO}} && $0.665$ & $0.455$ & $0.387$ && $0.855$ & $0.819$ & $0.794$ && $0.935$ & $0.923$ & $0.907$ && $0.577$ & $0.22$ & $0.187$ && $\mathbf{0.957}$ & $\mathbf{0.951}$ & $\mathbf{0.94}$ && $0.951$ & $0.943$ & $0.931$ && $0.892$ & $0.771$ & $0.719$ \\[-5pt] 
&&& $(0.045)$ & $(0.074)$ & $(0.038)$ && $(0.025)$ & $(0.03)$ & $(0.027)$ && $(0.005)$ & $(0.007)$ & $(0.009)$ && $(0.09)$ & $(0.229)$ & $(0.162)$ && $\mathbf{(0.003)}$ & $\mathbf{(0.002)}$ & $\mathbf{(0.003)}$ && $(0.017)$ & $(0.019)$ & $(0.021)$ && $(0.042)$ & $(0.027)$ & $(0.036)$ \\
& \multirow{2}{*}{\textit{JO}} && $0.428$ & $0.177$ & $-0.174$ && $0.505$ & $0.285$ & $0.013$ && $0.662$ & $0.539$ & $0.385$ && $0.438$ & $0.14$ & $-0.163$ && $0.678$ & $0.587$ & $0.399$ && $0.678$ & $0.585$ & $0.401$ && $\mathbf{0.725}$ & $\mathbf{0.697}$ & $\mathbf{0.461}$ \\[-5pt] 
&&& $(0.061)$ & $(0.161)$ & $(0.102)$ && $(0.005)$ & $(0.082)$ & $(0.012)$ && $(0.006)$ & $(0.019)$ & $(0.036)$ && $(0.06)$ & $(0.228)$ & $(0.141)$ && $(0.016)$ & $(0.009)$ & $(0.046)$ && $(0.018)$ & $(0.069)$ & $(0.013)$ && $\mathbf{(0.044)}$ & $\mathbf{(0.092)}$ & $\mathbf{(0.068)}$ \\
& \multirow{2}{*}{\itshape DS} && $0.582$ & $0.295$ & $0.262$ && $0.717$ & $0.61$ & $0.517$ && $0.723$ & $0.62$ & $\mathbf{0.531}$ && $0.533$ & $0.143$ & $0.083$ && $0.717$ & $0.605$ & $0.526$ && $0.718$ & $0.611$ & $0.523$ && $\mathbf{0.735}$ & $\mathbf{0.693}$ & $0.491$ \\[-5pt] 
&&& $(0.05)$ & $(0.175)$ & $(0.089)$ && $(0.01)$ & $(0.02)$ & $(0.016)$ && $(0.015)$ & $(0.031)$ & $\mathbf{(0.021)}$ && $(0.058)$ & $(0.247)$ & $(0.144)$ && $(0.003)$ & $(0.006)$ & $(0.004)$ && $(0.02)$ & $(0.038)$ & $(0.035)$ && $\mathbf{(0.013)}$ & $\mathbf{(0.016)}$ & $(0.041)$ \\[8pt]

\multirow{6}{*}{medium} & \multirow{2}{*}{\textit{RO}} && $0.643$ & $0.444$ & $0.475$ && $0.878$ & $0.836$ & $0.803$ && $0.944$ & $0.928$ & $0.911$ && $0.523$ & $0.086$ & $0.088$ && $0.957$ & $0.95$ & $0.94$ && $0.965$ & $0.953$ & $0.941$ && $\mathbf{0.975}$ & $\mathbf{0.955}$ & $\mathbf{0.944}$ \\[-5pt] 
&&& $(0.014)$ & $(0.034)$ & $(0.018)$ && $(0.013)$ & $(0.014)$ & $(0.018)$ && $(0.007)$ & $(0.008)$ & $(0.01)$ && $(0.041)$ & $(0.149)$ & $(0.153)$ && $(0.003)$ & $(0.003)$ & $(0.004)$ && $(0.006)$ & $(0.011)$ & $(0.013)$ && $\mathbf{(0.002)}$ & $\mathbf{(0.005)}$ & $\mathbf{(0.007)}$ \\
& \multirow{2}{*}{\textit{JO}} && $0.473$ & $0.006$ & $-0.133$ && $0.488$ & $0.376$ & $-0.023$ && $0.652$ & $0.558$ & $0.335$ && $0.463$ & $0.021$ & $-0.103$ && $0.675$ & $0.586$ & $0.388$ && $0.697$ & $0.655$ & $0.405$ && $\mathbf{0.867}$ & $\mathbf{0.855}$ & $\mathbf{0.743}$ \\[-5pt] 
&&& $(0.038)$ & $(0.01)$ & $(0.108)$ && $(0.029)$ & $(0.021)$ & $(0.061)$ && $(0.003)$ & $(0.002)$ & $(0.007)$ && $(0.064)$ & $(0.036)$ & $(0.178)$ && $(0.013)$ & $(0.01)$ & $(0.035)$ && $(0.013)$ & $(0.011)$ & $(0.028)$ && $\mathbf{(0.019)}$ & $\mathbf{(0.025)}$ & $\mathbf{(0.033)}$ \\
& \multirow{2}{*}{\itshape DS} && $0.535$ & $0.131$ & $0.19$ && $0.735$ & $0.646$ & $0.544$ && $0.743$ & $0.659$ & $0.559$ && $0.5$ & $0.0$ & $0.0$ && $0.715$ & $0.601$ & $0.523$ && $0.743$ & $0.659$ & $0.559$ && $\mathbf{0.773}$ & $\mathbf{0.712}$ & $\mathbf{0.604}$ \\[-5pt]
&&& $(0.009)$ & $(0.03)$ & $(0.024)$ && $(0.009)$ & $(0.017)$ & $(0.012)$ && $(0.003)$ & $(0.005)$ & $(0.005)$ && $(0.0)$ & $(0.0)$ & $(0.0)$ && $(0.0)$ & $(0.0)$ & $(0.0)$ && $(0.008)$ & $(0.014)$ & $(0.012)$ && $\mathbf{(0.016)}$ & $\mathbf{(0.024)}$ & $\mathbf{(0.029)}$ \\[8pt]

\multirow{6}{*}{large} & \multirow{2}{*}{\textit{RO}} && $0.641$ & $0.439$ & $0.482$ && $0.871$ & $0.834$ & $0.805$ && $0.944$ & $0.928$ & $0.911$ && $0.523$ & $0.085$ & $0.089$ && $0.956$ & $0.951$ & $0.94$ && $0.967$ & $0.958$ & $0.948$ && $\mathbf{0.976}$ & $\mathbf{0.959}$ & $\mathbf{0.949}$ \\[-5pt] 
&&& $(0.012)$ & $(0.029)$ & $(0.025)$ && $(0.014)$ & $(0.015)$ & $(0.016)$ && $(0.007)$ & $(0.008)$ & $(0.01)$ && $(0.04)$ & $(0.147)$ & $(0.153)$ && $(0.001)$ & $(0.003)$ & $(0.003)$ && $(0.004)$ & $(0.008)$ & $(0.01)$ && $\mathbf{(0.002)}$ & $\mathbf{(0.002)}$ & $\mathbf{(0.003)}$ \\
& \multirow{2}{*}{\textit{JO}} && $0.495$ & $0.0$ & $-0.071$ && $0.5$ & $0.352$ & $0.0$ && $0.652$ & $0.558$ & $0.335$ && $0.465$ & $0.021$ & $-0.099$ && $0.687$ & $0.589$ & $0.424$ && $0.72$ & $0.671$ & $0.462$ && $\mathbf{0.903}$ & $\mathbf{0.896}$ & $\mathbf{0.812}$ \\[-5pt]
&&& $(0.0)$ & $(0.0)$ & $(0.0)$ && $(0.0)$ & $(0.04)$ & $(0.0)$ && $(0.003)$ & $(0.002)$ & $(0.007)$ && $(0.061)$ & $(0.036)$ & $(0.172)$ && $(0.006)$ & $(0.014)$ & $(0.007)$ && $(0.023)$ & $(0.022)$ & $(0.052)$ && $\mathbf{(0.051)}$ & $\mathbf{(0.059)}$ & $\mathbf{(0.095)}$ \\
& \multirow{2}{*}{\itshape DS} && $0.53$ & $0.113$ & $0.176$ && $0.733$ & $0.641$ & $0.544$ && $0.743$ & $0.659$ & $0.559$ && $0.5$ & $0.0$ & $0.0$ && $0.715$ & $0.601$ & $0.523$ && $0.743$ & $0.659$ & $0.559$ && $\mathbf{0.777}$ & $\mathbf{0.717}$ & $\mathbf{0.61}$ \\[-5pt]
&&& $(0.0)$ & $(0.0)$ & $(0.0)$ && $(0.012)$ & $(0.021)$ & $(0.018)$ && $(0.003)$ & $(0.005)$ & $(0.005)$ && $(0.0)$ & $(0.0)$ & $(0.0)$ && $(0.0)$ & $(0.0)$ & $(0.0)$ && $(0.003)$ & $(0.005)$ & $(0.005)$ && $\mathbf{(0.01)}$ & $\mathbf{(0.018)}$ & $\mathbf{(0.016)}$\\

\bottomrule
\end{tabular}
}
\caption{\textbf{Experimental results}. Extension of Table~\ref{tab:results}. Balanced Accuracy (BA), $F_1$-score and Matthews correlation coefficient (MCC) for the Random Forest, Siamese Neural Network, and for the 5 single-distance Decision Trees, all trained with datasets of different sizes (small, medium, large) and tested on the three test sets discussed in Section~\ref{sec:training_and_test_sets}, namely randomly orderd ($S_\text{test}^\text{RO}$), JW-ordered ($S_\text{test}^\text{JO}$) and domain shifted ($S_\text{test}^\text{DS}$). Mean values and corresponding standard deviation (in brackets) computed via stratified $k$-fold cross-validation approach are reported for all metrics. Contrary to the 5 baseline models, the Random Forest and the Siamese Neural Network performances improve with the size of the training set. Moreover, the Siamese Neural Network trained on medium and large datasets outperforms all other approaches. For each of the three metrics, bold figures are row-wise maximum values.}
\label{tab:complete_results}
\end{center}
\end{table}
\end{landscape}
\restoregeometry
%

\end{document}